%% file: ex_arxiv.tex
\nonstopmode

\newif\ifAppendixEnabled
\newif\ifResultsEnlargedEnabled
\newif\ifAppendixExEnabled
\newif\ifAppendixGammaEnabled
\newif\ifAppendixArgmaxEnabled
\newif\ifAppendixHyperparametersEnabled
\newif\ifAppendixExperimentalSetupEnabled
\newif\ifAppendixRUSEnabled

\AppendixEnabledtrue
\ResultsEnlargedEnabledtrue
\AppendixExEnabledtrue
\AppendixGammaEnabledtrue
\AppendixArgmaxEnabledtrue
\AppendixHyperparametersEnabledtrue
\AppendixExperimentalSetupEnabledtrue
\AppendixRUSEnabledtrue

\relax
\RequirePackage{amsmath}
\documentclass{llncs}
\usepackage[bb=boondox]{mathalfa}
\usepackage{float} 
\usepackage{times}
\usepackage{helvet}
\usepackage{courier}
\usepackage[hyphens]{url}
\usepackage{graphicx}
\usepackage{graphicx}
\usepackage{amsmath}
\usepackage[utf8]{inputenc}
\usepackage{subcaption}
\usepackage{amssymb}
\usepackage{bbm}
\usepackage{tabularx}
\usepackage{listings}
\usepackage[linesnumbered,ruled,lined]{algorithm2e}

\DeclareMathOperator*{\argmin}{\arg\!\min}

\usepackage{colortbl}
\definecolor{myred}{rgb}{0.8,0,0}
\definecolor{mygreen}{rgb}{0,0.6,0}
\definecolor{myblue}{rgb}{0,0,0.7}
\newcommand{\todo}[1]{}

\newcommand{\ours}{{\sc PBCS}\xspace}

\title{\ours: Efficient Exploration and Exploitation Using a Synergy between Reinforcement Learning and Motion Planning}
\author{Guillaume Matheron\inst{1}\orcidID{0000-0001-6530-8784} \and
        Nicolas Perrin\inst{1}\orcidID{0000-0003-2358-2915} \and 
        Olivier Sigaud\inst{1}\orcidID{0000-0002-8544-0229}}
\authorrunning{G. Matheron et al.}
\institute{Sorbonne Université, CNRS, Institut des Systèmes Intelligents et de Robotique, ISIR,\\ F-75005 Paris, France}

\begin{document}

\maketitle

\begin{abstract}
The exploration-exploitation trade-off is at the heart of reinforcement learning (RL). However, most continuous control benchmarks used in recent RL research only require local exploration. This led to the development of algorithms that have basic exploration capabilities, and behave poorly in benchmarks that require more versatile exploration.
For instance, as demonstrated in our empirical study, state-of-the-art RL algorithms such as DDPG and TD3 are unable to steer a point mass in even small 2D mazes.
In this paper, we propose a new algorithm called "Plan, Backplay, Chain Skills" (\ours) that combines motion planning and reinforcement learning to solve hard exploration environments.
In a first phase, a motion planning algorithm is used to find a single good trajectory, then an RL algorithm is trained using a curriculum derived from the trajectory, by combining a variant of the Backplay algorithm and skill chaining.
We show that this method outperforms state-of-the-art RL algorithms in 2D maze environments of various sizes, and is able to improve on the trajectory obtained by the motion planning phase.
\end{abstract}

\section*{Introduction}

Reinforcement Learning (RL) algorithms have been used successfully to optimize policies for both discrete and continuous control problems with high dimensionality~\cite{mnih_playing_2013,lillicrap_continuous_2015}, but fall short when trying to solve difficult exploration problems~\cite{van_hasselt_deep_2018,achiam_towards_2019,schaul_prioritized_2015}. On the other hand, motion planning (MP) algorithms such as RRT~\cite{lavalle_rapidly-exploring_1998} are able to efficiently explore in large cluttered environments but, instead of trained policies, they output trajectories that cannot be used directly for closed loop control.

In this paper, we consider environments that present a hard exploration problem with a sparse reward. In this context, a \emph{good trajectory} is one that reaches a state with a positive reward, and we say that an environment is \emph{solved} when a controller is able to reliably reach a rewarded state. We illustrate our approach with 2D continuous action mazes as they facilitate the visual examination of the results, but we believe that this approach can be beneficial to many robotics problems.

If one wants to obtain closed loop controllers for hard exploration problems, a simple approach is to first use an MP algorithm to find a single good trajectory $\tau$, then optimize and robustify it using RL. However, using $\tau$ as a stepping stone for an RL algorithm is not straightforward. In this article, we propose \ours, an approach that fits the framework of Go-Explore~\cite{ecoffet_go-explore_2019}, and is based on the Backplay algorithm~\cite{resnick_backplay_2018} and skill chaining~\cite{konidaris_skill_2009,konidaris_constructing_2010}. We show that this approach greatly outperforms both DDPG~\cite{lillicrap_continuous_2015} and TD3~\cite{fujimoto_addressing_2018} on continuous control problems in 2D mazes, as well as approaches that use Backplay but no skill chaining.

\ours has two successive phases.
First, the environment is explored until a single good trajectory is found.
Then this trajectory is used to create a curriculum for training DDPG.
More precisely, \ours progressively increases the difficulty through a backplay process which gradually moves the starting point of the environment backwards along the trajectory resulting from exploration.
Unfortunately, this process has its own issues, and DDPG becomes unstable in long training sessions.
Calling upon a skill chaining approach, we use the fact that even if Backplay eventually fails, it is still able to solve some subset of the problem. Therefore, a partial policy is saved, and the reminder of the problem is solved recursively until the full environment can be solved reliably.

In this article, we contribute an extension of the Go-Explore framework to continuous control environments, a new way to combine a variant of the Backplay algorithm with skill chaining, and a new state-space exploration algorithm.

\section{Related Work}
\label{sec:related}

Many works have tried to incorporate better exploration mechanisms in RL, with various approaches.

\paragraph{Encouraging exploration of novel states}
The authors of~\cite{tang_exploration_2016} use a count-based method to penalize states that have already been visited, while the method proposed by~\cite{benureau_behavioral_2016} reuses actions that have provided diverse results in the past.
Some methods try to choose policies that are both efficient and novel~\cite{pugh_confronting_2015,cully_quality_2017,pugh_quality_2016,erickson_survivability_2009}, while some use novelty as the only target, entirely removing the need for rewards~\cite{eysenbach_diversity_2018,knepper_path_2009}.
The authors of \cite{stadie_incentivizing_2015,burda_exploration_2018,pathak_curiosity-driven_2017} train a forward model and use the unexpectedness of the environment step as a proxy for novelty, which is encouraged through reward shaping.
Some approaches try to either estimate the uncertainty of value estimates~\cite{osband_deep_2016}, or learn bounds on the value function~\cite{ciosek_better_2019}.
All these solutions try to integrate an exploration component within RL algorithms, while our approach separates exploration and exploitation into two successive phases, as in~\cite{colas_gep-pg_2018}.


\paragraph{Using additional information about the environment}
Usually in RL, the agent can only learn about the environment through interactions.
However, when additional information about the task at hand is provided, other methods are available.
This information can take the form of expert demonstrations~\cite{salimans_learning_2018,hosu_playing_2016,resnick_backplay_2018,konidaris_constructing_2010,fournier_clic_2019,nair_overcoming_2018,paine_making_2019}, or having access to a single rewarded state~\cite{florensa_reverse_2018}.
When a full representation of the environment is known, RL can still be valuable to handle the dynamics of the problem: PRM-RL~\cite{faust_prm-rl_2018} and RL-RRT~\cite{chiang_rl-rrt_2019} use RL as reachability estimators during a motion planning process.

\paragraph{Building on the Go-Explore framework}
To our knowledge, the closest approach to ours is the Go-Explore~\cite{ecoffet_go-explore_2019} framework, but in contrast to \ours, Go-Explore is applied to discrete problems such as Atari benchmarks.
In a first phase, a single valid trajectory is computed using an ad-hoc exploration algorithm.
In a second phase, a learning from demonstration (LfD) algorithm is used to imitate and improve upon this trajectory.
Go-Explore uses Backplay~\cite{resnick_backplay_2018,salimans_learning_2018} as the LfD algorithm, with Proximal Policy Optimization (PPO)~\cite{schulman_proximal_2017} as policy optimization method.
Similar to Backplay, the authors of~\cite{goyal_recall_2019} have proposed Recall Traces, a process in which a backtracking model is used to generate a collection of trajectories reaching the goal.

The authors of~\cite{morere_reinforcement_2020} present an approach that is similar to ours, and also fits the framework of Go-Explore. In phase 1, they use a guided variant of RRT, and in phase 2 they use a learning from demonstration algorithm based on TRPO.
Similarly, \ours follows the same two phases as Go-Explore, with major changes to both phases. In the first phase, our exploration process is adapted to continuous control environments by using a different binning method, and different criteria for choosing the state to reset to. In the second phase, a variant of Backplay is integrated with DDPG instead of PPO, and seamlessly integrated with a skill chaining strategy and reward shaping.

The Backplay algorithm in \ours is a deterministic variant of the one proposed in~\cite{resnick_backplay_2018}. In the original Backplay algorithm, the starting point of each policy is chosen randomly from a subset of the trajectory, but in our variant the starting point is deterministic: the last state of the trajectory is used until the performance of DDPG converges (more details are presented in Sect.~\ref{sec:backplay}), then the previous state is chosen, and so on until the full trajectory has been exploited.

\paragraph{Skill chaining}
The process of \emph{skill chaining} was explored in different contexts by several research papers.
The authors of~\cite{konidaris_skill_2009} present an algorithm that incrementally learns a set of skills using classifiers to identify changepoints, while the method proposed in~\cite{konidaris_constructing_2010} builds a skill tree from demonstration trajectories, and automatically detects changepoints using statistics on the value function.
To our knowledge, our approach is the first to use Backplay to build a skill chain. We believe that it is more reliable and minimizes the number of changepoints because the position of changepoints is decided using data from the RL algorithm that trains the policies involved in each skill.

\section{Background}
\label{sec:background}

Our work is an extension of the Go-Explore algorithm. In this section, we summarize the main concepts of our approach.

\paragraph{Reset-anywhere.} Our work makes heavy use of the ability to reset an environment to any state. The use of this primitive is relatively uncommon in RL, because it is not always readily available, especially in real-world robotics problems. However, it can be invaluable to speed up exploration of large state spaces. It was used in the context of Atari games by~\cite{hosu_playing_2016}, proposed in \cite{schulman_trust_2015} as {\scshape vine}, and gained popularity with~\cite{salimans_learning_2018}.

\paragraph{Sparse rewards.} Most traditional RL benchmarks only require very local exploration, and have smooth rewards guiding them towards the right behavior. Thus sparse rewards problems are especially hard for RL algorithms: the agent has to discover without any external signal a long sequence of actions leading to the reward. Most methods that have been used to help with this issue require prior environment-specific knowledge \cite{riedmiller_learning_2018}.

\paragraph{Maze environments.}
Lower-dimension environments such as cliff walk~\cite{sutton_reinforcement_2018} are often used to demonstrate fundamental properties of RL algorithms, and testing in these environments occasionally reveals fundamental flaws~\cite{matheron_problem_2019}.
We deliberately chose to test our approach on 2D maze environments because they are hard exploration problems, and because reward shaping behaves very poorly in such environments, creating many local optima.
Our results in Section~\ref{sec:results} show that state-of-the-art algorithms such as DDPG and TD3 fail to solve even very simple mazes.

\paragraph{DDPG.}
Deep Deterministic Policy Gradient (DDPG) is a continuous action actor-critic algorithm using a deterministic actor that performs well on many control tasks~\cite{lillicrap_continuous_2015}. However, DDPG suffers from several sources of instability.
Our maze environments fit the analysis made by the authors of~\cite{penedones_temporal_2018}, according to whom the critic approximator may "leak" Q-value across walls of discontinuous environments.
With a slightly different approach,~\cite{fujimoto_off-policy_2018} suggests that extrapolation error may cause DDPG to over-estimate the value of states that have never been visited or are unreachable, causing instability.
More generally, the authors of~\cite{sutton_reinforcement_2018} formalize the concept of "deadly triad", according to which algorithms that combine function approximation, bootstrapping updates and off-policy are prone to diverge. Even if the deadly triad is generally studied in the context of the DQN algorithm~\cite{mnih_playing_2013}, these studies could also apply to DDPG.
Finally, the authors of~\cite{matheron_problem_2019} show that DDPG can fail even in trivial environments, when the reward is not found quickly enough by the built-in exploration of DDPG.

\section{Methods}

\begin{figure}
    \centering
    \includegraphics[width=.6\columnwidth]{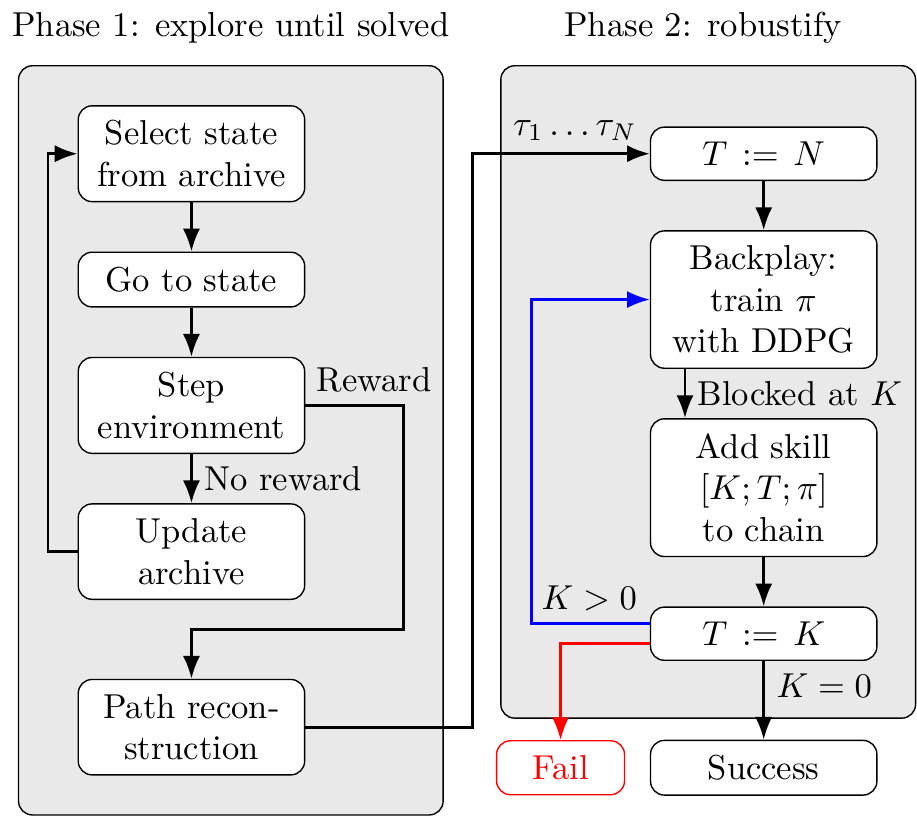}
    \caption{Overview of \ours. The red path is only used when testing the algorithm without skill chaining, otherwise the blue path is used.}
    \label{fig:algo}
\end{figure}

Figure~\ref{fig:algo} describes \ours. The algorithm is split in two successive phases, mirroring the Go-Explore framework. In a first phase, the environment is incrementally explored until a single rewarded state is found.
In a second phase, a single trajectory provides a list of starting points, that are used to train DDPG on increasingly difficult portions of the full environment. Each time the problem becomes too difficult and DDPG starts to fail, training stops, and the trained agent is recorded as a local skill. Training then resumes for the next skill, with a new target state.
This loop generates a set of skills that can then be chained together to create a controller that reaches the target reliably.

\subsection*{Notations}

\paragraph{State neighborhood.} For any state $\tau_i\in S$, and $\epsilon>0$, we define $B_\epsilon(\tau_i)$ as the closed ball of radius $\epsilon$ centered around $\tau_i$. Formally, this is the set $\left\{s\in S\mid d(s,\tau_i)\leq \epsilon\right\}$ where $d$ is the L2 distance.

\paragraph{Skill chaining.} Skill chaining consists in splitting a complex task into simpler sub-tasks that are each governed by a different policy. Complex tasks can then be solved by executing each policy sequentially.

Formally, each task $T_i$ has an activation condition $A_i~\subset~S$, and a policy $\pi_i: S\rightarrow A$. A \emph{task chain} is a list of tasks $T_0 \ldots T_n$, which can be executed sequentially: the actor uses $\pi_0$ until the state of the system reaches a state $s\in A_1$, then it uses $\pi_1$, and so on until the end of the episode (which can be triggered by reaching either a terminal state or a predetermined maximum number of steps).

\subsection{Phase~1: Explore until Solved}
\label{sec:phase1}

In phase~1, \ours explores to find a single path that obtains a non-zero reward in the environment.
\ifAppendixExEnabled
  This exploration phase is summarized in this section, and implementation details are available in Appendix~\ref{sec:ex}.
\fi
An archive keeps track of all the visited states. In this archive, states $s$ are grouped in square state-space bins. A state-counter $c_s$ is attached to each state, and a bin-counter $c_b$ is attached to each bin. All counters are initialized to $0$.

The algorithm proceeds in 5 steps, as depicted in \figurename~\ref{fig:algo}:

\begin{enumerate}
  \item \textbf{Select state from archive.}\enspace
To select a state, the non-empty bin with the lowest counter is first selected, then from all the states in this bin, the state with the lowest counter is selected. Both the bin and state counters are then incremented.

  \item \textbf{Go to state.}\enspace
The environment is reset to the selected state. This assumes the existence of a "reset-anywhere" primitive, which can be made available in simulated environments.

  \item \textbf{Step environment.}\enspace
A single environment step is performed, with a random action.

  \item \textbf{Update archive.}\enspace
The newly-reached state is added to the archive if not already present.

  \item \textbf{Termination of phase~1.}\enspace
As soon as the reward is reached, the archive is used to reconstruct the sequence of states that led the agent from its initial state to the reward. This sequence $\tau_0 \ldots \tau_N$ is passed on to phase 2.
\end{enumerate}

This process can be seen as a random walk with a constraint on the maximum distance between two states: in the beginning, a single trajectory is explored until it reaches a state that is too close to an already-visited state. When this happens, a random visited state is selected as the starting point of a new random walk.
Another interpretation of this process is the construction of a set of states with uniform spatial density. Under this view, the number of states in each cell is used as a proxy for the spatial density of the distribution.

\subsection{Phase 2: Robustify}
\label{sec:phase2}

Phase 2 of \ours learns a controller from the trajectory obtained in phase~1.

\subsubsection{Skill Chaining.}
\label{sec:skill_chaining}

\begin{algorithm}[htb]
  \DontPrintSemicolon
  \SetKwFunction{FBackplay}{Backplay}
  \SetKwInOut{Input}{Input}
  \SetKwInOut{Output}{Output}
  \Input{
    $\tau_0\ldots \tau_N$ the output of phase~1\;
  }
  \Output{
    $\pi_0\ldots\pi_n$ a chain of policies with activation sets $A_0\ldots A_n$
  }
  $T = N$ \;
  $n = 0$ \;
  \While{$T>0$}{
    $\pi_n,T = \FBackplay(\tau_0\ldots\tau_T)$ \;
    $A_n = B_\epsilon(\tau_T)$ \;
    $n = n + 1$
  }
  Reverse lists $\pi_0\ldots\pi_n$ and $A_0\ldots A_n$ \;
  \caption{Phase 2 of \ours}
  \label{lst:skill_chaining}
\end{algorithm}

Algorithm~\ref{lst:skill_chaining} presents the skill chaining process. It uses the \texttt{Backplay} function, that takes as input a trajectory $\tau_0\ldots\tau_T$, and returns a policy $\pi$ and an index $K<T$ such that running policy $\pi$ repeatedly on a state from $B_\epsilon(\tau_K)$ always leads to a state in $B_\epsilon(\tau_T)$.
The main loop builds a chain of skills that roughly follows trajectory $\tau$, but is able to improve upon it. Specifically, activation sets $A_n$ are centered around points of $\tau$ but policies $\pi_n$ are constructed using a generic RL algorithm that optimizes the path between two activation sets.
The list of skills is then reversed, because it was constructed backwards.

\subsubsection{Backplay.}
\label{sec:backplay}

The Backplay algorithm was originally proposed in~\cite{resnick_backplay_2018}.
\ifAppendixRUSEnabled
  More details on the differences between this original algorithm and our variant are available in sections~\ref{sec:related} and~\ref{sec:pm}.
\else
  More details on the differences between this original algorithm and our variant are available in Section~\ref{sec:related}.
\fi

The \texttt{Backplay} function (Algorithm~\ref{lst:backplay}) takes as input a section $\tau_0 \ldots \tau_T$ of the trajectory obtained in phase~1, and returns a $(K,\pi)$ pair where $K$ is an index on trajectory $\tau$, and $\pi$ is a policy trained to reliably attain $B_\epsilon(\tau_T)$ from $B_\epsilon(\tau_K)$. 
The policy $\pi$ is trained using DDPG to reach $B_\epsilon(\tau_T)$ from starting point $B_\epsilon(\tau_K)$
\ifAppendixRUSEnabled
  \footnote{More details on why the starting point needs to be $B_\epsilon(\tau_K)$ instead of $\tau_K$ are available in Appendix~\ref{sec:pm}}
\fi
, where $K$ is initialized to $T-1$, and gradually decremented in the main loop.

At each iteration, the algorithm evaluates the feasibility of a skill with target $B_\epsilon(\tau_T)$, policy $\pi$ and activation set $B_\epsilon(\tau_K)$. If the measured performance is $100\%$ without any training (line~\ref{l:ifp1}), the current skill is saved and the starting point is decremented.
Otherwise, a training loop is executed until performance stabilizes (line~\ref{l:train}). This is performed by running Algorithm~\ref{lst:training} repeatedly until no improvement over the maximum performance is observed $\alpha$ times in a row. We ran our experiments with $\alpha=10$ %
\ifAppendixHyperparametersEnabled %
, and a more in-depth discussion of hyperparameters is available in Appendix~\ref{sec:hyperparameters}
\fi.

Then the performance of the skill is measured again (line~\ref{l:test2}), and three cases are handled:

\begin{itemize}
    \item \textbf{The skill is always successful (line~\ref{l:p21}).} The current skill is saved and the index of the starting point is decremented.
    \item \textbf{The skill is never successful (line~\ref{l:p20}).} The last successful skill is returned.
    \item \textbf{The skill is sometimes successful.} The current skill is not saved, and the index of the starting point is decremented. In our maze environment, this happens when $B_\epsilon(\tau_K)$ overlaps a wall: in this case some states of $B_\epsilon(\tau_K)$ cannot reach the target no matter the policy.
\end{itemize}

\begin{algorithm}[htbp]
  \DontPrintSemicolon
  \SetKwFunction{FTrain}{Train}
  \SetKwProg{Fn}{Function}{:}{}
  \SetKwInOut{Input}{Input}
  \SetKwInOut{Output}{Output}
  \Input{
        $(\tau_0\ldots\tau_T)$ a state-space trajectory
        }
  \Output{  $\pi_s$ a trained policy \newline
            $K_s$ the index of the starting point of the policy
        }
  $K = T - 1$ \;
  Initialize a DDPG architecture with policy $\pi$ \;
  \While{$K>0$}{
    Test performance of $\pi$ between $B_\epsilon(\tau_K)$ and $B_\epsilon(\tau_T)$ over $\beta$ episodes \;
    \eIf{performance $= 100\%$}{  \label{l:ifp1}
      $\pi_s = \pi$, $K_s = K$
    }{
      Run \FTrain (Algorithm~\ref{lst:training}) repeatedly until performance stabilizes.\; \label{l:train}
      Test performance of $\pi$ between $B_\epsilon(\tau_K)$ and $B_\epsilon(\tau_T)$ over $\beta$ episodes \;\label{l:test2}
      \If{performance $= 100\%$}{ \label{l:p21}
        $\pi_s = \pi$, $K_s = K$
      }
      \If{performance $= 0\%$ and $K_s$ exists}{  \label{l:p20}
        \textbf{return} $(K_s, \pi_s)$
      }
    }
    $K = K - 1$ \;
  }
  \textbf{return} $(K_s, \pi_s)$

  \caption{The Backplay algorithm}
  \label{lst:backplay}
\end{algorithm}

\subsubsection{Reward Shaping.}
\label{sec:reward_shaping}

With reward shaping, we bypass the reward function of the environment, and train DDPG to reach any state $\tau_T$. We chose to use the method proposed by~\cite{ng_policy_1999}: we define a potential function in Equation~\eqref{eq:potential}, where $d(s,A_i)$ is the L2 distance between $s$ and the center of $A_i$. We then define our shaped reward in Equation~\eqref{eq:reward}.

\begin{subequations}
  \begin{alignat}{1}
    \Phi(s) &= \frac{1}{d(s, A_i)} \label{eq:potential} \\
    R_\text{shaped}(s,a,s') &= \begin{cases}10&\text{if } s\in A_i \\ \Phi(s') - \Phi(s)&\text{otherwise.}\end{cases} \label{eq:reward}
\end{alignat}
\end{subequations}

\begin{algorithm}[htbp]
  \DontPrintSemicolon
  \SetKwInOut{Input}{Input}
  \SetKwInOut{Output}{Output}
  \Input{
        $\tau_K$ the source state \newline
        $\tau_T$ the target state
        }
  \Output{
        The average performance $p$
        }
    n = 0 \;
    \For{$i=1\ldots \beta$}{
      $s \sim B_\epsilon(\tau_K)$ \;
      \For{$j=1\ldots \text{max\_steps}$}{
        $a = \pi(s) + \text{random noise}$ \;
        $s' = \text{step}(s,a)$          \nllabel{line:step_test}\;
        $r = \begin{cases}
          10&d(s',\tau_T)\leq\epsilon\\
          \frac{1}{d(s',\tau_T)} - \frac{1}{d(s,\tau_T)}&\text{otherwise}
          \end{cases}$\;
        DDPG.train$(s,a,s',r)$\;
        $s = s'$\;
        \If{$d(s',\tau_T) \leq \epsilon$}{
          n = n + 1\;
          \textbf{break}
        }
      }
    }
    $p=\frac{n}{\beta}$ \nllabel{l:trt2}
  \caption{Training process with reward shaping}
  \label{lst:training}
\end{algorithm}

Algorithm~\ref{lst:training} shows how this reward function is used in place of the environment reward. This training function runs $\beta$ episodes of up to \texttt{max\_steps} steps each, and returns the fraction of episodes that were able to reach the reward. $\beta$ is a hyperparameter that we set to 50 for our test%
\ifAppendixHyperparametersEnabled %
, and more details on this choice are available in Appendix~\ref{sec:hyperparameters}
\fi.

Importantly, reaching a performance of 100\% is not always possible, even with long training sessions, because the starting point is selected in $B_\epsilon(\tau_K)$, and some of these states may be inside obstacles for instance.

\section{Experimental Results}
\label{sec:results}

\begin{figure*}
    \captionof{table}{Results of various algorithms on maze environments. For each test, the number of environment steps performed is displayed with a red background when the policy was not able to reach the target, and a green one when training was successful. \newline
    In "Vanilla" experiments, the red paths represent the whole area explored by the RL algorithm. In "Backplay" experiments, the trajectory computed in phase~1 is displayed in red, and the "robustified" policy or policy chain is displayed in green. Activation sets $A_i$ are displayed as purple circles.
      \ifResultsEnlargedEnabled
        Enlarged images are presented in Fig.~\ref{fig:results_big}.
      \fi
    }
    \label{fig:results}
    \input{results.tex}
\end{figure*}

We perform experiments in continuous maze environments of various sizes. For a maze of size $N$, the state-space is the position of a point mass in $[0,N]^2$ and the action describes the speed of the point mass, in $[-0.1, 0.1]^2$. Therefore, the step function is simply $s'=s+a$, unless the $[s,s']$ segment intersects a wall. The only reward is $-1$ when hitting a wall and $1$ when the target area is reached.
\ifAppendixExperimentalSetupEnabled
A more formal definition of the environment is available in Appendix~\ref{sec:setup}.
\fi

Our results are presented in Table~\ref{fig:results}. We first tested standard RL algorithms (DDPG and TD3), then \ours, but without skill chaining (this was done by replacing the blue branch with the red branch in \figurename~\ref{fig:algo}). When the full algorithm would add a new skill to the skill chain and continue training, this variant stops and fails. These results are presented in column "\ours without skill chaining".
Finally, the full version of \ours with skill chaining is able to solve complex mazes up to $15 \times 15$ cells, by chaining several intermediate skills.

\section{Discussion of Results}

As expected, standard RL algorithms (DDPG and TD3) were unable to solve all but the simplest mazes. These algorithms have no mechanism for state-space exploration other than uniform noise added to their policies during rollouts. Therefore, in the best-case scenario they perform a random walk and, in the worst-case scenario, their actors may actively hinder exploration.

More surprisingly, \ours without skill chaining is still unable to reliably
\ifAppendixGammaEnabled
\footnote{We tested \ours without skill chaining with different seeds on small mazes, these results are presented in Appendix~\ref{sec:gamma}}
\fi
solve mazes larger than $2 \times 2$. Although phase~1 always succeeds in finding a feasible trajectory $\tau$, the robustification phase fails relatively early. We attribute these failures to well-known limitations of DDPG exposed in Section~\ref{sec:background}.
\ifAppendixGammaEnabled
We found that the success rate of \ours without skill chaining was very dependent on the discount rate $\gamma$, which we discuss in Appendix~\ref{sec:gamma}.
\fi

The full version of \ours with skill chaining is able to overcome these issues by limiting the length of training sessions of DDPG, and is able to solve complex mazes up to $7 \times 7$, by chaining several intermediate skills.

\section{Conclusion}

The authors of Go-Explore identified state-space exploration as a fundamental difficulty on two Atari benchmarks. We believe that this difficulty is also present in many continuous control problems, especially in high-dimension environments. We have shown that the \ours algorithm can solve these hard exploration, continuous control environments by combining a motion planning process with reinforcement learning and skill chaining. Further developments should focus on testing these hybrid approaches on higher dimensional environments that present difficult exploration challenges together with difficult local control, such as the Ant-Maze MuJoCo benchmark~\cite{tassa_deepmind_2018}, and developing methods that use heuristics suited to continuous control in the exploration process, such as Quality-Diversity approaches \cite{pugh_quality_2016}.

\section{Acknowledgements}

This work was partially supported by the French National Research Agency (ANR), Project ANR-18-CE33-0005 HUSKI.

\bibliographystyle{splncs04}
\bibliography{ex_arxiv}{}

\ifAppendixEnabled
  
  \clearpage
  \begin{center}
    \textbf{\Huge Supplemental Materials}
  \end{center}
  \setcounter{equation}{0}
  \setcounter{figure}{0}
  \setcounter{table}{0}
  \setcounter{section}{0}
  \setcounter{subsection}{0}
  \setcounter{page}{1}
  \makeatletter
  \renewcommand{\theequation}{S\arabic{equation}}
  \renewcommand{\thefigure}{S\arabic{figure}}
  \renewcommand{\thetable}{S\arabic{table}}
  \renewcommand{\thesection}{S\arabic{section}}
\fi

\ifAppendixExEnabled
  \section{Phase~1: Explore Until Solved}
  \label{sec:ex}
  \begin{algorithm}[htbp]
    \DontPrintSemicolon
    \SetKwFunction{KwTrue}{True}
    \SetKwFunction{KwNot}{Not}
    \SetKwInOut{Input}{Input}
    \SetKwInOut{Output}{Output}
    \Input{$s_0 \in S$ the initial environment state \newline
          step $: S \times A \rightarrow S \times \mathbb{R} \times \mathbb{B}$
            the environment step function\newline
          iterations $ \in \mathbb{N}$ the number of samples to accumulate \newline
          Bin $: F \rightarrow \mathbb{N}$ a binning function
          }
    \Output{transitions $ \subseteq S \times A \times \mathbb{R} \times \mathbb{B} \times S$ the set of explored transitions \newline
          }
    transitions = $\varnothing$                                  \;
    $b_0$ = Bin($s_0$)                               \nllabel{line:init}\;
    bin\_usage[$b_0$] = 0                                \;
    states\_in\_bin[$b_0$] = \{$s_0$\}                             \;
    state\_usage[$s_0$] = 0                              \;
    B = B $\cup$ \{$b_0$\}                                \nllabel{line:initEnd}\;
    \While{$\left|\text{\upshape transitions}\right| < $ \upshape iterations}{
      chosen\_bin = $\displaystyle \argmin_{b \in B}$ bin\_usage[b]             \nllabel{line:binSelection}\;
      chosen\_state = $\displaystyle \argmin_{s \in states\_in\_bin[chosen\_bin]}$ state\_usage[chosen\_state]   \nllabel{line:stateSelection} \nllabel{line:selectionEnd}\;
      action = random\_action() \;
      s',reward,terminal = step(chosen\_state, action)          \nllabel{line:step}\;
  
      bin\_usage[chosen\_bin] ++\;
      state\_usage[chosen\_state] ++\;
      transitions = transitions $\cup$ (chosen\_state,action,reward,terminal,s') \;
  
      \If{\KwNot \upshape terminal}{
          state\_usage[s'] = 0                          \nllabel{line:insert} \;
          b' = Bin(s')                                 \;
          \eIf{b' $\in$ B}{
              states\_in\_bin[b'] = states\_in\_bin[b'] $\cup$ \{s'\}      \;
          }
          {
              B = B $\cup$ \{b'\}                       \;
              bin\_usage[b'] = 0                       \;
              states\_in\_bin[b'] = \{s'\}                       \nllabel{line:insertEnd}
          }
      }
    }
    \caption{Exploration algorithm}
    \label{lst:ex}
  \end{algorithm}
  
  Our proposed phase~1 exploration algorithm maintains a pool of states, which initially contains only the start state. At each step, a selection process described below is used to select a state $s$ from the pool (lines~\ref{line:binSelection} to~\ref{line:selectionEnd} in Algorithm~\ref{lst:ex}). A random action $a$ is then chosen, and the environment is used to compute a single step from state $s$ using action $a$ (line~\ref{line:step}). If the resulting state $s'$ is non-terminal, then it is added to the pool (lines~\ref{line:insert} to~\ref{line:insertEnd}). This process is repeated as long as necessary.
  
  \paragraph{State selection} The pool of states stored by the algorithm is divided into square bins of size $0.05$. During the state selection process (lines~\ref{line:binSelection} to~\ref{line:selectionEnd}), the least-chosen bin is selected (on line~\ref{line:binSelection}), then the least-chosen state from this bin is selected (line~\ref{line:stateSelection}). In both cases, when several states or bins are tied in the \lstinline|argmin| operation, one is selected uniformly randomly from the set of all tied elements.
\fi

\ifAppendixHyperparametersEnabled
  \section{Choice of \ours Hyperparameters}
  \label{sec:hyperparameters}
  
  \ours uses three hyperparameters $\alpha$, $\beta$ and $\epsilon$.
  
  The parameter $\alpha$ represents the number of consecutive non-improvements of the training performance required to assume training is finished. In our experiments, this value was set to $10$, and we summarize here what can be expected if this parameter is set too high or too low.
  
  \begin{itemize}
      \item Setting $\alpha$ too high results in longer training sessions, in which the policy keeps being trained despite being already successful. The time and sample performance of \ours is impacted, but the algorithm should still be able to build a policy chain.
      \item Setting $\alpha$ too low may cause training to stop early. In benign cases, the policy is simply sub-optimal, but in some cases this may lead to the creation of many changepoints, and prevent \ours from improving at all upon the phase~1 trajectory. If the activation conditions overlap too much, \ours may output a skill chain that is unable to navigate the environment.
  \end{itemize}
  
  The parameter $\beta$ represents the number of samples used to evaluate the performance of a skill. In our experiments, this value was set to $50$.
  
  \begin{itemize}
      \item Setting $\beta$ too high would increase the time and sample complexity of \ours, but would not impact the output.
      \item The main risk of setting $\beta$ too low is that \ours may incorrectly compute that a skill has a performance of $100\%$. If this skill is then selected, the output skill chain may be unable to navigate the environment.
  \end{itemize}
  
  The parameter $\epsilon$ corresponds to the radius of the targets used during skill chaining.
\fi

\ifAppendixGammaEnabled
  \section{Choice of Discount Factor $\gamma$}
  \label{sec:gamma}
  
  The discount factor $\gamma$ is usually considered to be a parameter of the environment and not the RL algorithm. It controls the decay that is applied when evaluated the contributions of future rewards to a present choice.  In our experiments, we tested two values of $\gamma$, that are $\gamma=0.9$ and $\gamma=0.99$.
  
  DDPG uses a neural network $\hat{Q}$ in order to estimate the state-action value function $Q^\pi(s,a)$ of the current policy $\pi$. In the case of deterministic environments, the state-action value function is recursively defined as $Q^\pi(s,a) = R(s,a,s') + \gamma Q(s',\pi(a))$, where $s' = \text{step}(s,a)$.
  
  Therefore, reaching a sparse reward of value $1$ after $n$ steps with no reward carries a discounted value of $\gamma^n$. This implies that rewards that are reached only after many steps have very little impact on the shape of $Q$. For instance, with $\gamma=0.9$ and $n=50$, $\gamma^n\approx 0.05$. This effectively reduces the magnitude of the training signal used by the actor update of DDPG, reducing the speed of actor updates the farther from the reward. Evidence of this is presented in Sect.~\ref{sec:argmax}.
  
  With this consideration, it seems that choosing $\gamma$ very close to $1$ solves the problem of exponential decay of the training signal. However, high $\gamma$ values present their own challenges. The state-action critic approximator $\hat{Q}$ used by DDPG is trained on $(s,a,r,s')$ tuples stored in an experience replay buffer, but as with any continuous approximator, it generalizes the training data to nearby $(s,a)$ couples.
  
  In environments with positive rewards, $\hat{Q}$ can over-estimate the value of states: for instance in maze environments, the learned value can be generalized incorrectly and "leak" through walls.
  
  This mechanism is usually counter-balanced by the fact that over-estimated $Q(s,a)$ values can then be lowered. For instance, in our maze environments, hitting a wall generates a training tuple with $s'=s$ and $r=0$. The update rule of DDPG applied to this tuple yields: $Q(s,a)\leftarrow Q(s,a) (1+c(\gamma-1))$ where $c$ is the critic learning rate. Therefore, the closer $\gamma$ is to $1$, the slower over-estimated values will be corrected.
  
  In smaller mazes, our experiments show that reducing gamma increases the performance of \ours without skill chaining (Fig.~\ref{fig:influence_gamma}). \todo{Interpretation}
\fi

\ifAppendixArgmaxEnabled
  \subsection{Replacing the Actor Update of DDPG}
  \label{sec:argmax}
  
  We claim that the lower reward signal obtained with $\gamma$ values close to $1$ affect the actor update of DDPG. We can test this claim by using a variation of DDPG proposed by the authors of~\cite{matheron_problem_2019}: we replace the actor update of DDPG with a brute-force approach that, for each sampled state $s$, computes $\max_a \hat{Q}(s,a)$ using uniform sampling. The performance of this variant is presented in Fig.~\ref{fig:influence_gamma} with green and red bars. \todo{discuss results}
  
  \begin{figure}
    \centering
    \includegraphics{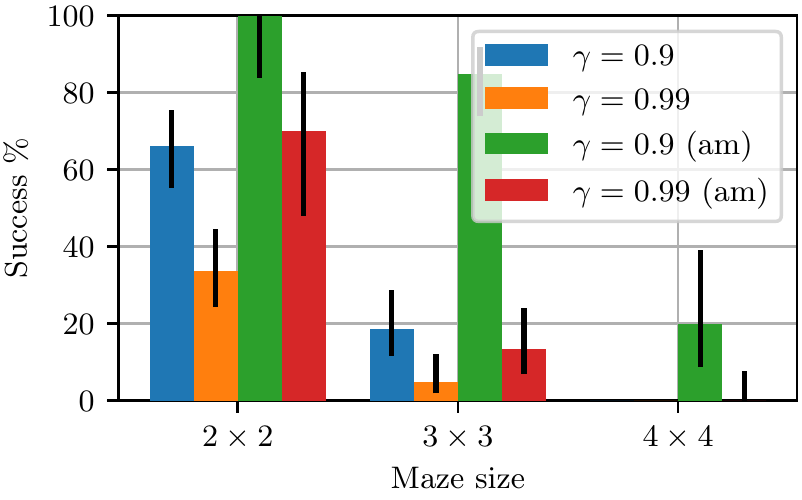}
    \captionof{figure}{Success rate of \ours without skill chaining, depending on $\gamma$.
      Bars marked with \emph{(am)} use the variant of DDPG presented in Sect.~\ref{sec:argmax}.
      Error bars are computed using Wilson score intervals~\cite{wilson_probable_1927}.}
    \label{fig:influence_gamma}
  \end{figure}
\fi

\ifAppendixExperimentalSetupEnabled
  \section{Experimental Setup}
  \label{sec:setup}
  
  Our experiments are conducted in maze environments of various sizes. A maze of size $N$ is described using the following Markov Decision Process:
  
  \begin{align*}
      S &= [0,N]\times [0,N] \\
      A &= [-0.1,0.1] \times[-0.1,0.1] \\
      R(s,a,s') &= \mathbb{1}_{\|s'-\text{target}\| < 0.2}
          - \mathbb{1}_{[s,s'] \text{ intersects a wall}}\\
      \text{step}(s,a) &= \begin{cases}s&\text{if } [s,s+a] \text{ intersects a wall}\\
          s+a&\text{otherwise.}\end{cases}
  \end{align*}
  
  The set of walls is constructed using a maze generation algorithm, and walls have a thickness of 0.1.
  
  The target position is $(N-.5,N-.5)$ when $N>2$. In mazes of size 2, the target position is $(.5,1.5)$.
\fi

\ifResultsEnlargedEnabled
  \section{Enlarged Results}
  
  A more detailed view of the results of \ours on mazes of different sizes is presented in Fig.~\ref{fig:results}.
  
  \begin{figure}[ht]
      \caption{Enlarged view of the results of \ours on mazes of different sizes. The trajectory computed in phase~1 is displayed in red, and the "robustified" policy chain is displayed in green. Activation sets $A_i$ are displayed as purple circles.}
      \label{fig:results_big}
      \centering
      \input{results_big.tex}
  \end{figure}
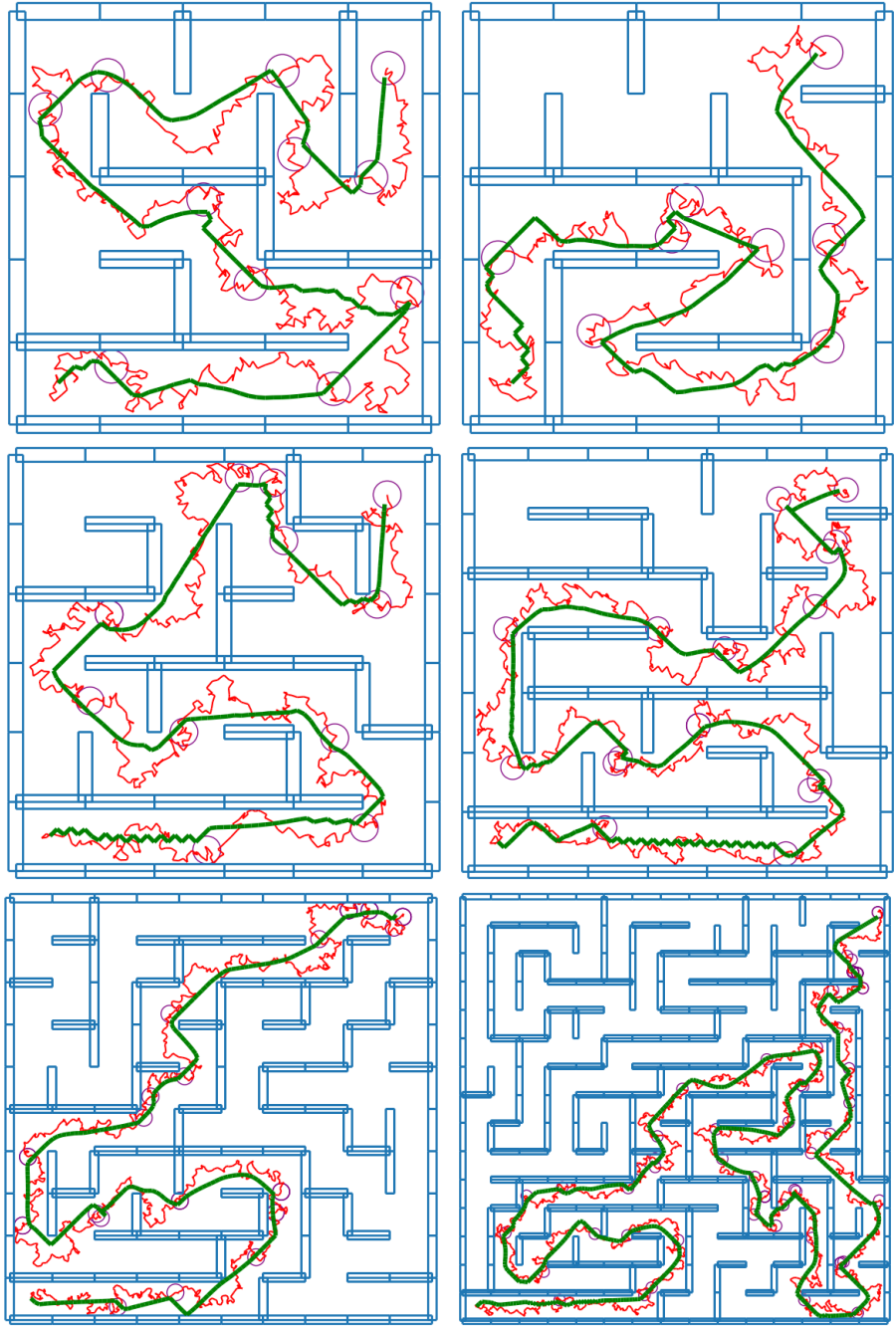
\fi

\ifAppendixRUSEnabled
  \section{Need for Resetting in Unseen States}
  \label{sec:pm}
  
  As a reminder, for the Backplay algorithm and our variant, a single trajectory $\tau_0\ldots\tau_T$ is provided, and training is performed by changing the starting point of the environment to various states.
  
  In the original Backplay algorithm, the environment is always reset to a visited state $\tau_K$, where $K$ is an index chosen randomly in a sliding window of $[0,T]$. The sliding window is controlled by hyperparameters, but the main idea is that in the early stages of training, states near $T$ are more likely to be selected, and in later stages, states near $0$ are more likely to be used.
  
  However, we found that this caused a major issue when combined with continuous control and the skill chaining process. With skill chaining, the algorithm creates a sequence of activation sets $(A_n)$, and a sequence of policies $(\pi_n)$ such that when the agent reaches a state in $A_n$, it switches to policy $\pi_n$. Each activation set $A_n$ is a ball of radius $\epsilon$ centered around a state $\tau_K$ for some K.
  
  The policy needs to be trained not only on portions of the environment that are increasingly long, it also needs to account for the uncertainty of its starting point. When executing the skill chain, the controller switched to policy $\pi_n$ as soon as the state reaches the activation set $A_n$, which is $B_\epsilon(\tau_K)$ for some K. Even if $A_n$ is relatively small, we found it caused systematic issues on maze environments, as presented in Fig.~\ref{fig:pmproblem}.
  
  \begin{figure}[p]
      \centering
      \includegraphics[width=.6\linewidth]{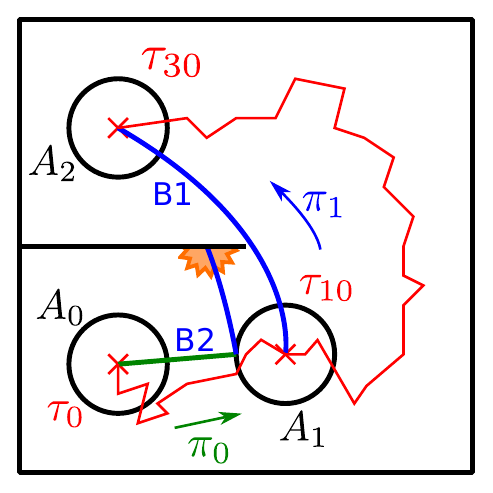}
      \caption{Policy $\pi_1$ was trained using starting points $\tau_{30}\ldots\tau_{10}$ without any added noise. Therefore, $\tau_{30}$ is reachable from $\tau_{10}$ using $\pi_1$ (trajectory $B2$), but not necessarily form any point in $A_1$. In maze environments, the optimal policy is usually close to walls, and provides little margin for perturbations.
      The trajectory $B2$ (that starts in green and ends in blue) results from the execution of the skill chain. The controller switches from $\pi_0$ to $\pi_1$ as soon as the agent reaches $A_1$, and then hits the wall (trajectory $B2$). This problem persists even when $\epsilon$ is reduced.}
      \label{fig:pmproblem}
  \end{figure}
  
  In our variant of the Backplay algorithm, we found it was necessary to train DDPG on starting points chosen randomly in $B_\epsilon(\tau_K)$, to ensure that the policy is trained correctly to solve a portion of the environment with any starting point in this volume.
  
  This also means that we need to reset the environment to unseen states, and can cause problems when these states are unreachable (in our maze examples this is usually because they are inside walls, but in higher dimensions we assume this could be more problematic).
  
  When possible, a solution would be to run the environment backwards from $\tau_K$ with random actions to generate these samples (while ensuring that they still lie within $B_\epsilon(\tau_K)$). Another solution, especially in high-dimension environments, would be to run the environment backwards for a fixed number of steps, and use a classifier to define the bounds of $A_n$, instead of using the L2 distance.
\fi

\end{document}

%% file: results.tex
\setlength\tabcolsep{0pt}
\newcommand{\bad}[1]{\colorbox{red}{#1}}
\newcommand{\good}[1]{\colorbox{green}{#1}}
\newcommand{\neutral}[1]{\colorbox{yellow}{#1}}
\newcolumntype{Y}{>{\centering\arraybackslash}X}
\begin{tabularx}{\textwidth}{ |Y|Y|Y|Y|  }
  \hline
    \multicolumn{2}{|c|}{Vanilla} %
  & \multicolumn{1}{c|}{\ours w/o skill chaining} %
  & \multicolumn{1}{c|}{\ours} %
  \\ \hline
  DDPG & TD3 & DDPG & DDPG \\ \hline

\includegraphics[trim=160px 80px 140px 80px,clip,width=2.5cm]{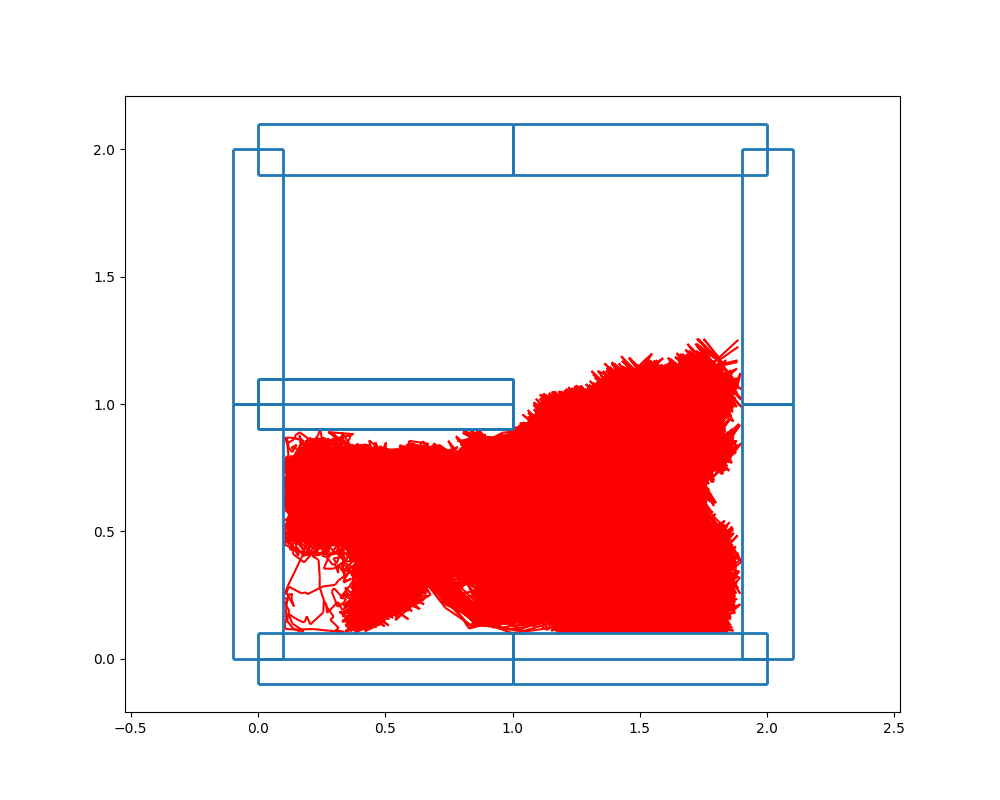} \bad{1M}& \includegraphics[trim=160px 80px 140px 80px,clip,width=2.5cm]{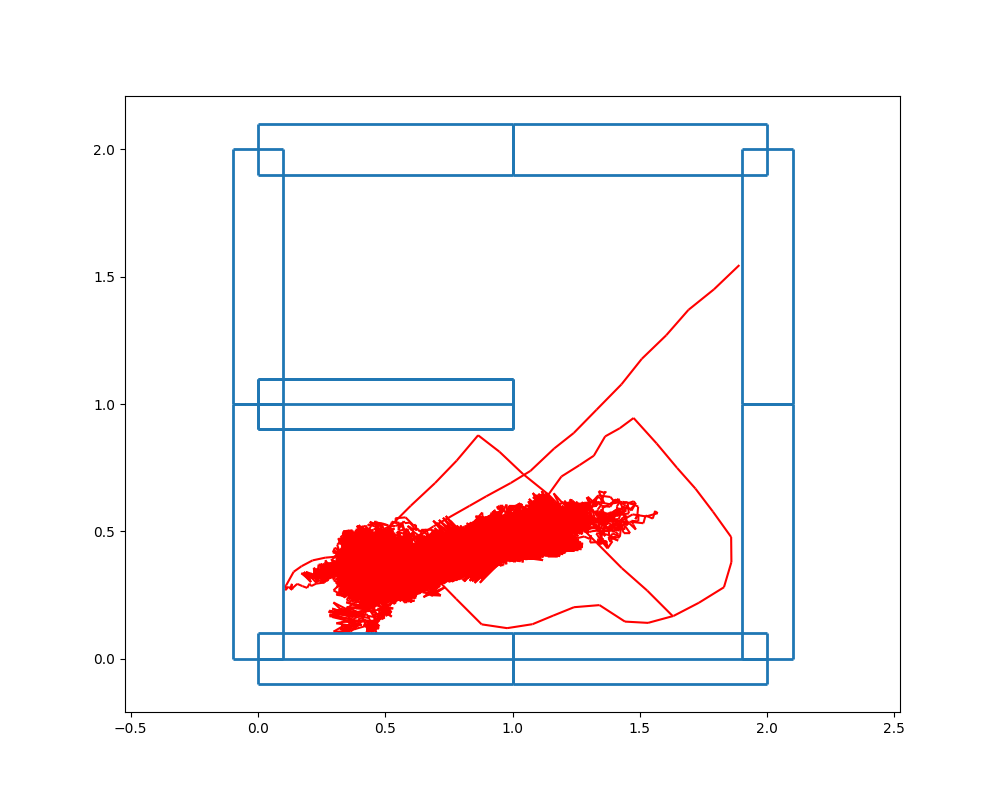} \bad{1M}& \includegraphics[trim=160px 80px 140px 80px,clip,width=2.5cm]{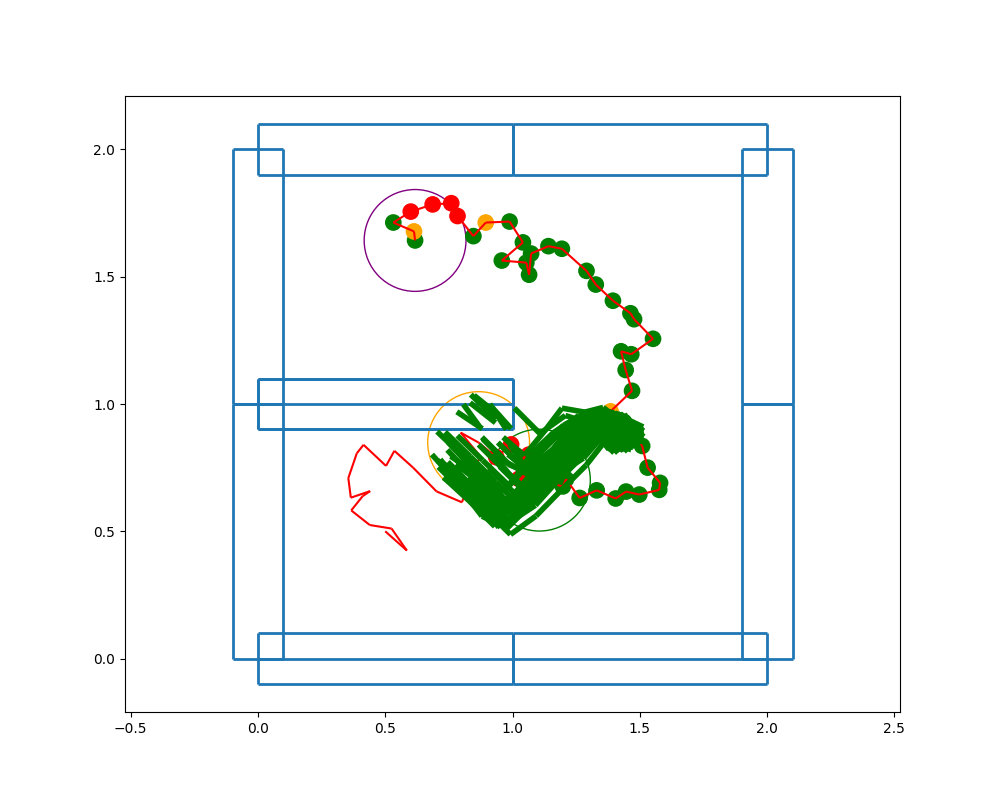} \bad{146k}& \includegraphics[trim=160px 80px 140px 80px,clip,width=2.5cm]{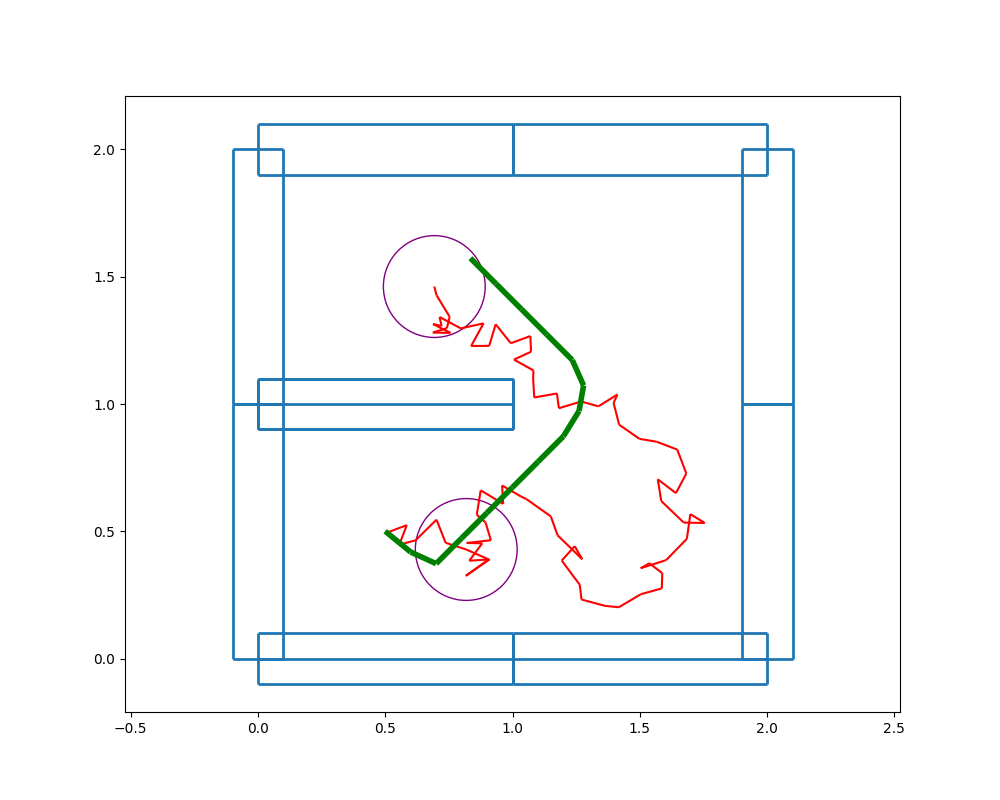} \good{321k}\\ \hline
\includegraphics[trim=160px 80px 140px 80px,clip,width=2.5cm]{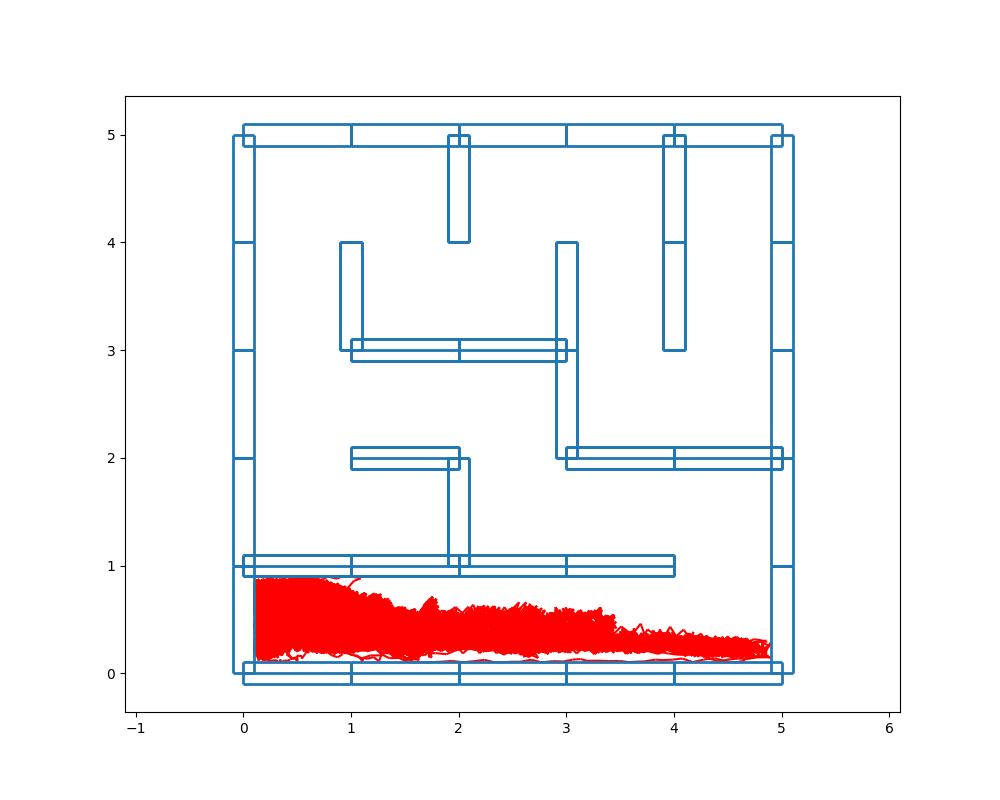} \bad{1M}& \includegraphics[trim=160px 80px 140px 80px,clip,width=2.5cm]{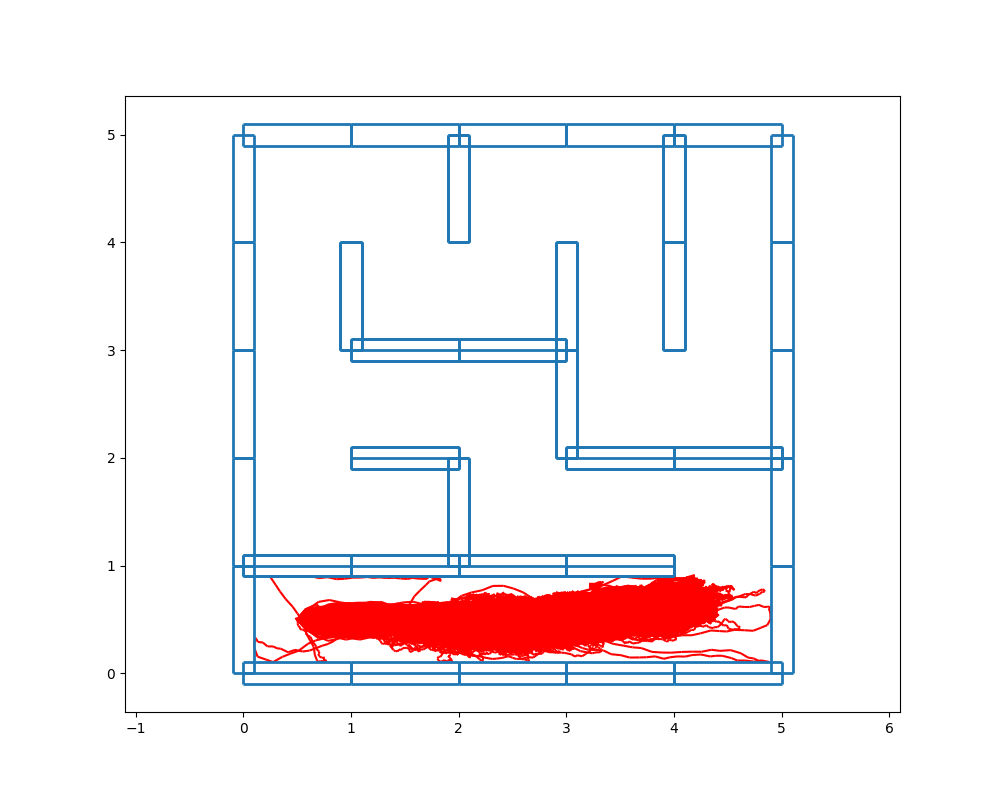} \bad{1M}& \includegraphics[trim=160px 80px 140px 80px,clip,width=2.5cm]{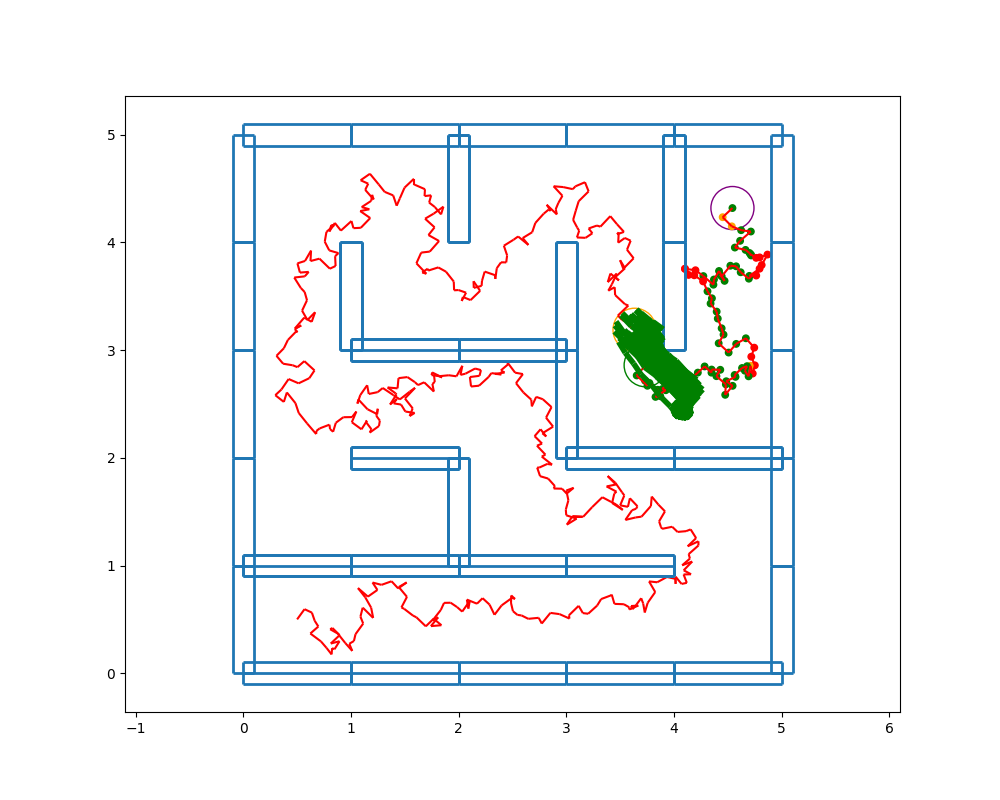} \bad{372k}& \includegraphics[trim=160px 80px 140px 80px,clip,width=2.5cm]{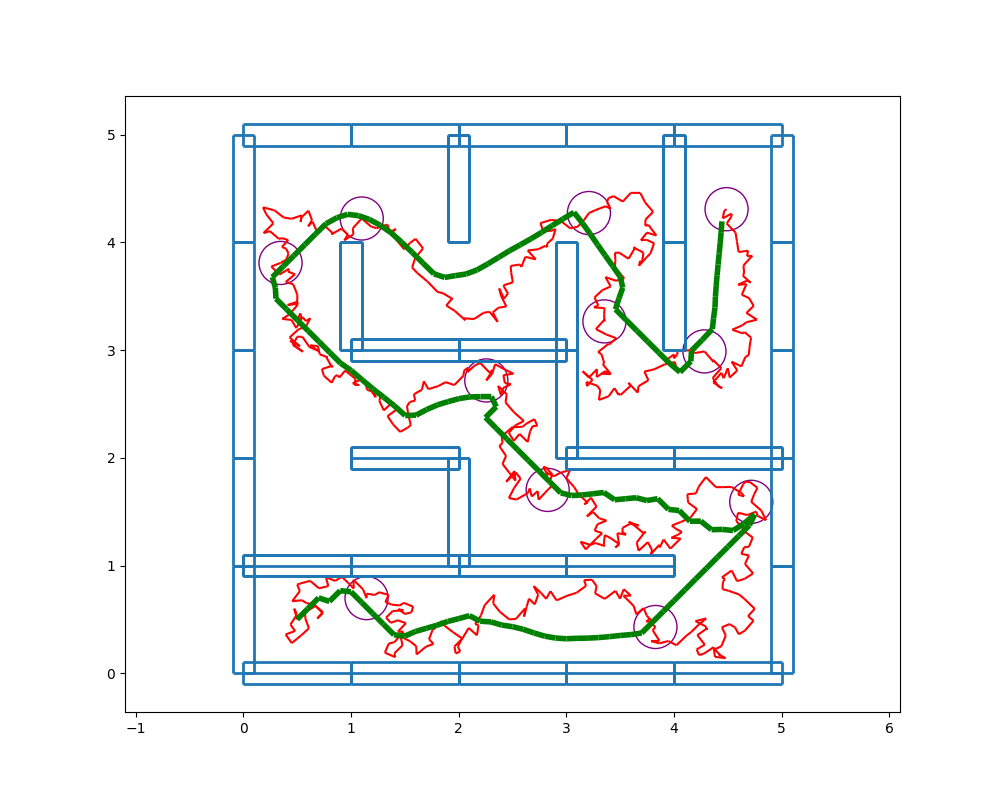} \good{5M}\\ \hline
\includegraphics[trim=160px 80px 140px 80px,clip,width=2.5cm]{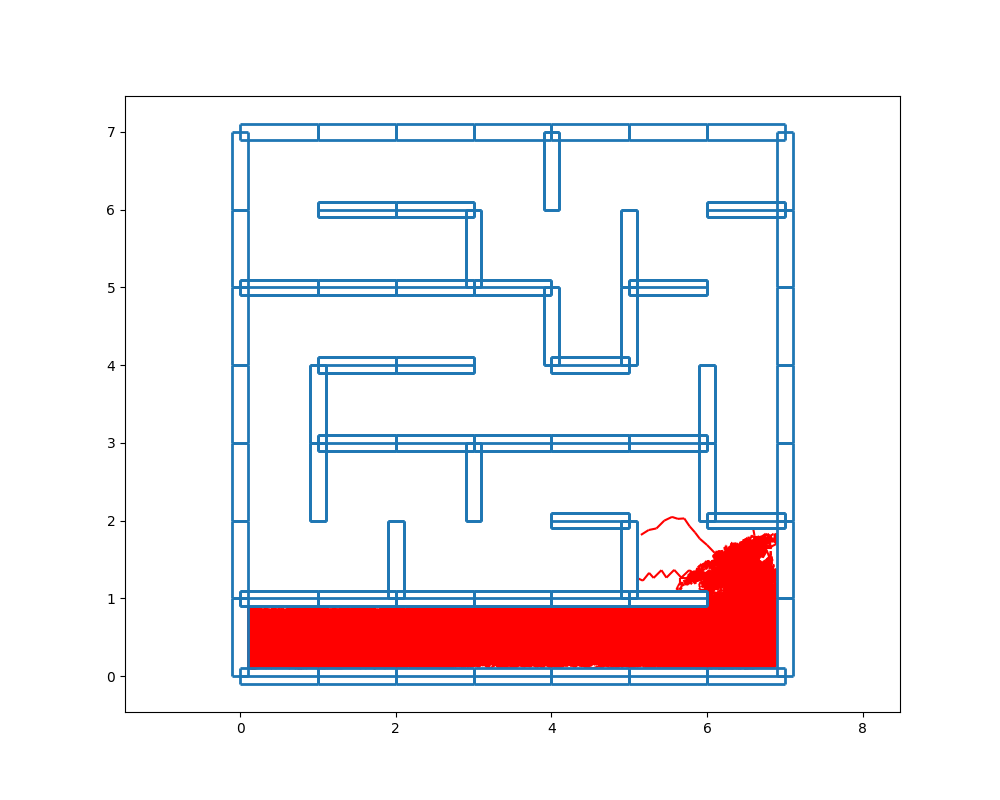} \bad{1M}& \includegraphics[trim=160px 80px 140px 80px,clip,width=2.5cm]{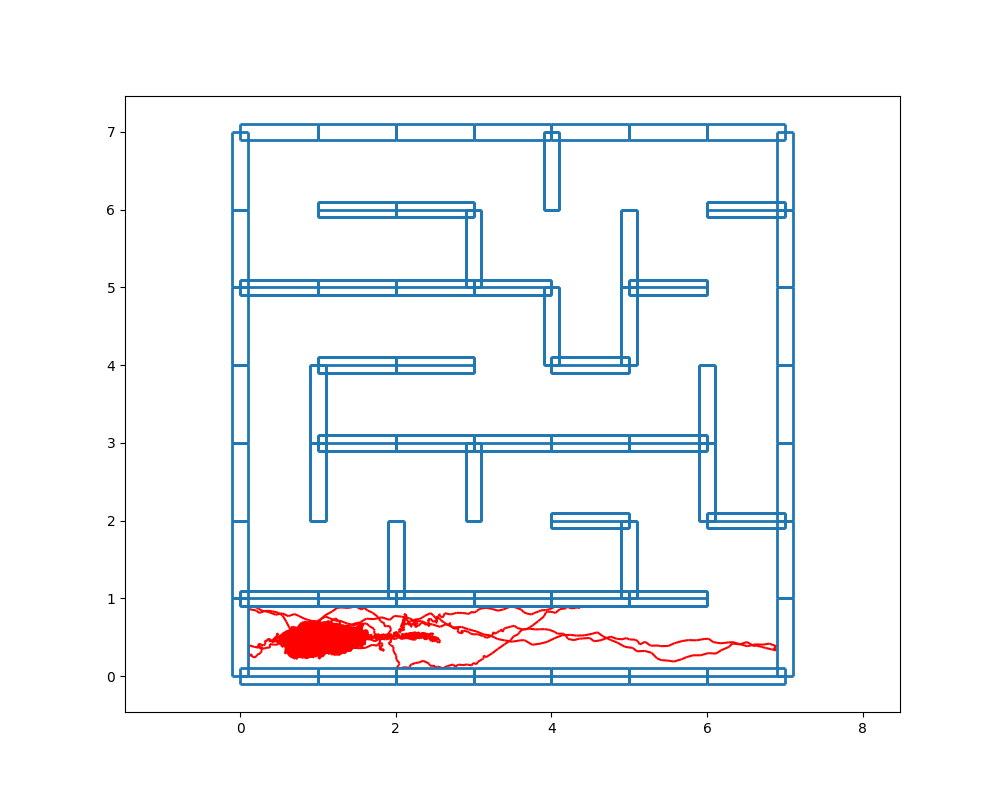} \bad{1M}& \includegraphics[trim=160px 80px 140px 80px,clip,width=2.5cm]{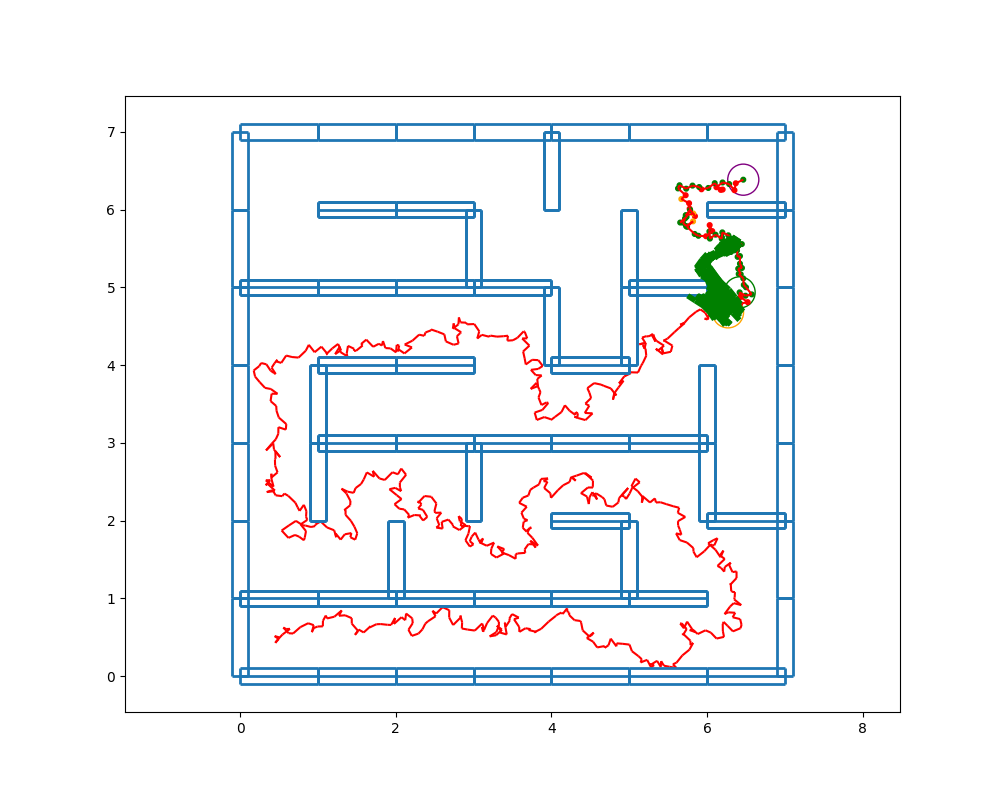} \bad{268k}& \includegraphics[trim=160px 80px 140px 80px,clip,width=2.5cm]{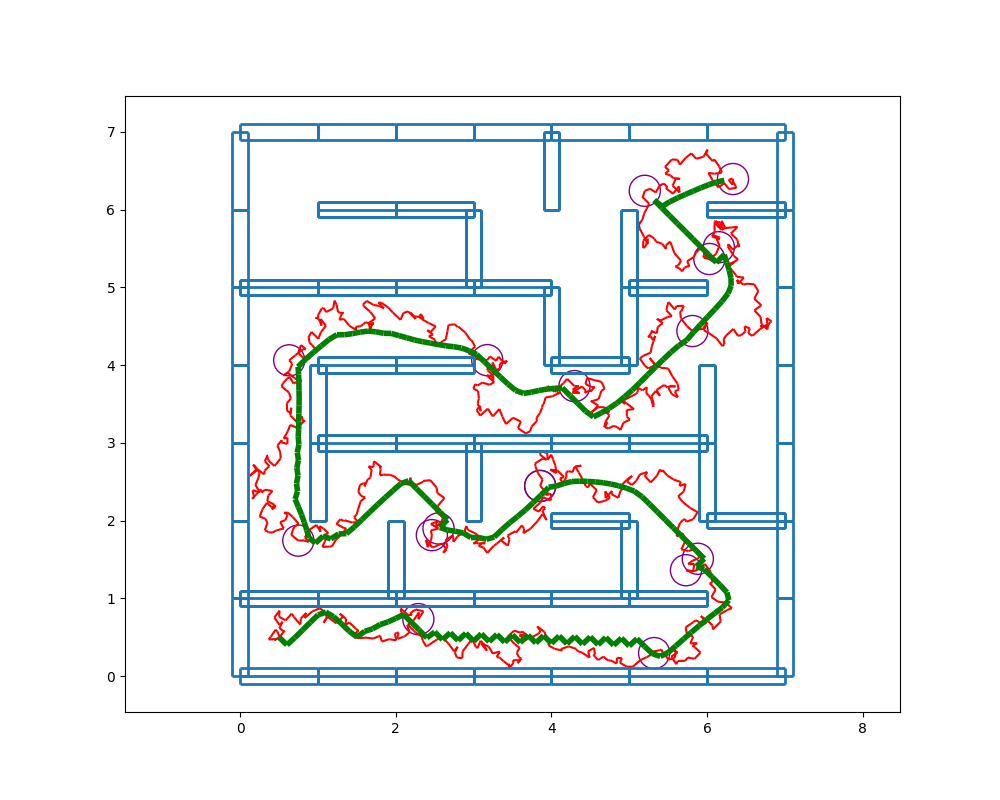} \good{6M}\\ \hline
\includegraphics[trim=160px 80px 140px 80px,clip,width=2.5cm]{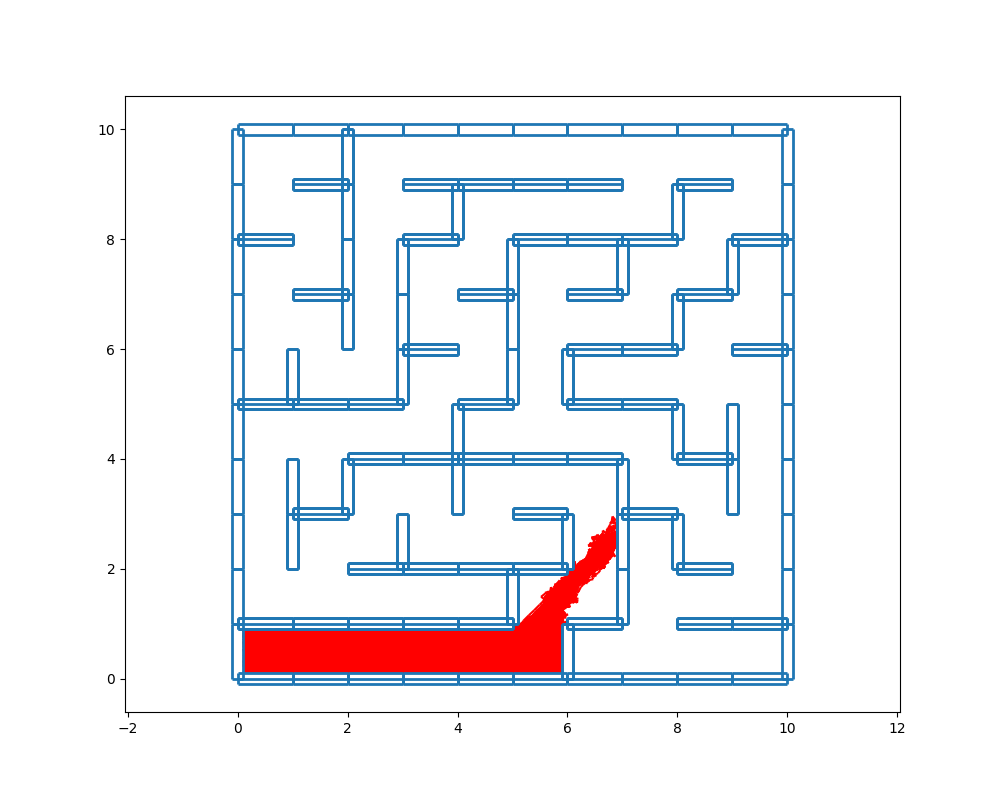} \bad{1M}& \includegraphics[trim=160px 80px 140px 80px,clip,width=2.5cm]{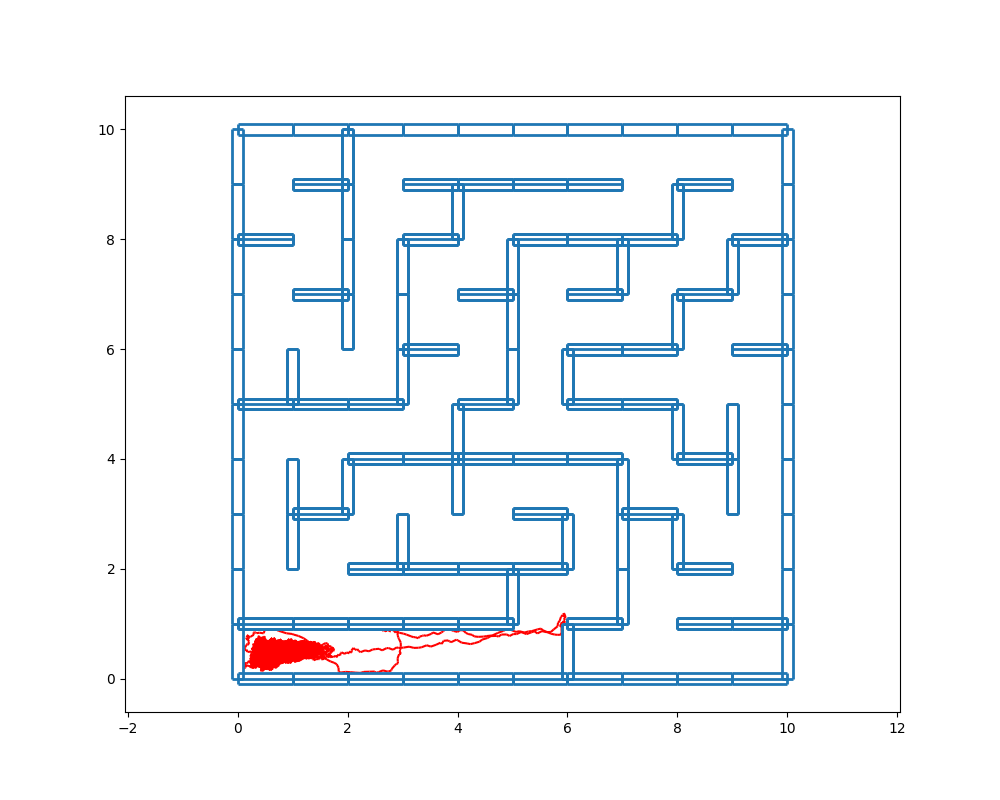} \bad{1M}& \includegraphics[trim=160px 80px 140px 80px,clip,width=2.5cm]{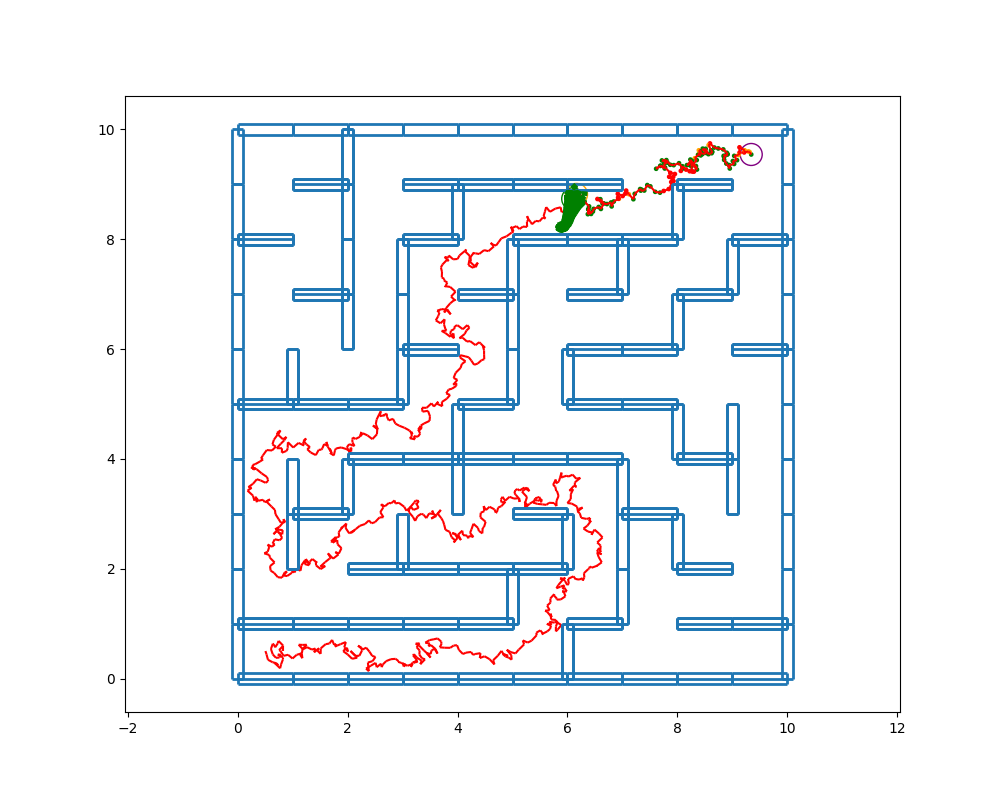} \bad{694k}& \includegraphics[trim=160px 80px 140px 80px,clip,width=2.5cm]{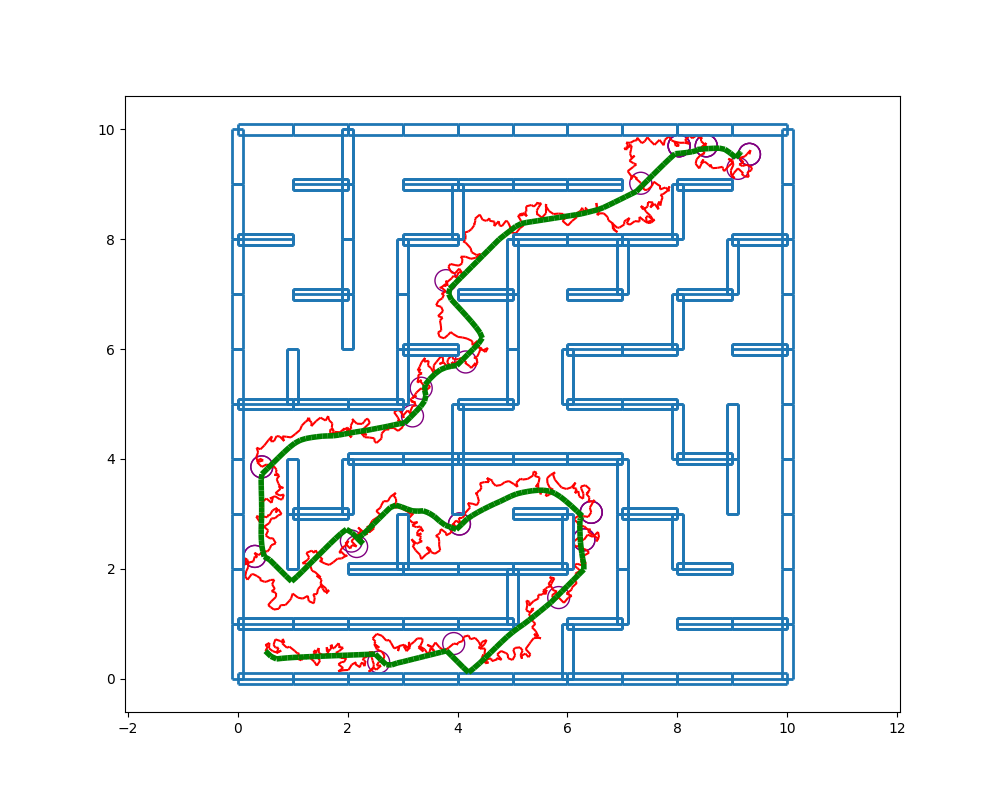} \good{8M}\\ \hline
\includegraphics[trim=160px 80px 140px 80px,clip,width=2.5cm]{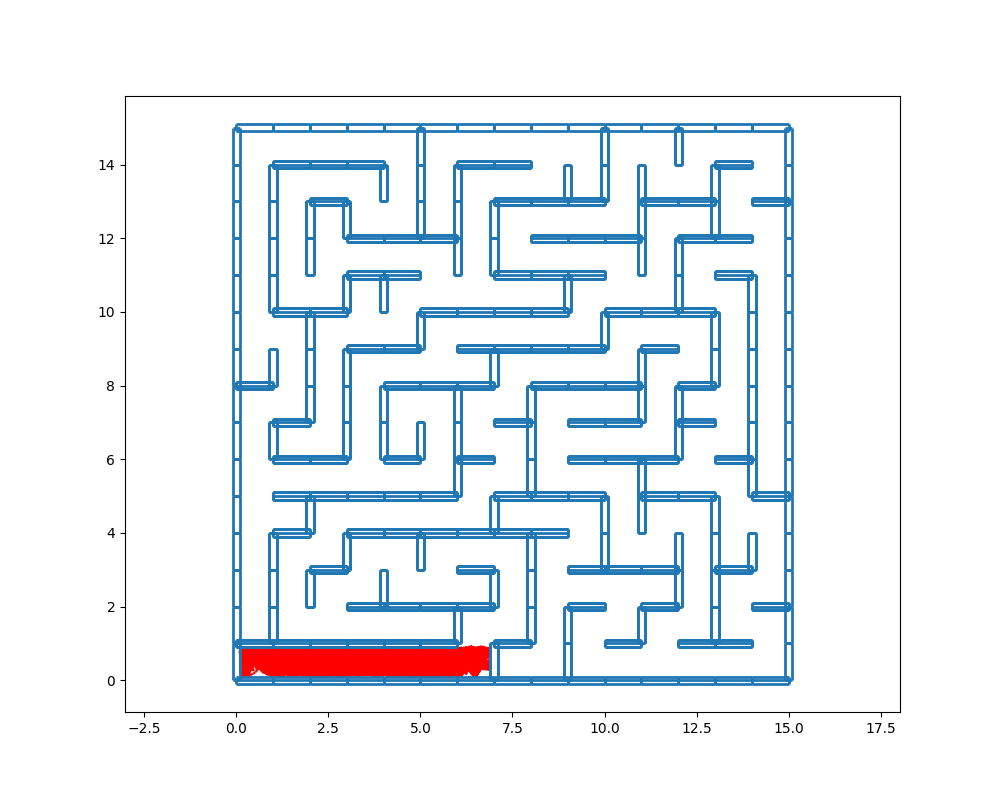} \bad{1M}& \includegraphics[trim=160px 80px 140px 80px,clip,width=2.5cm]{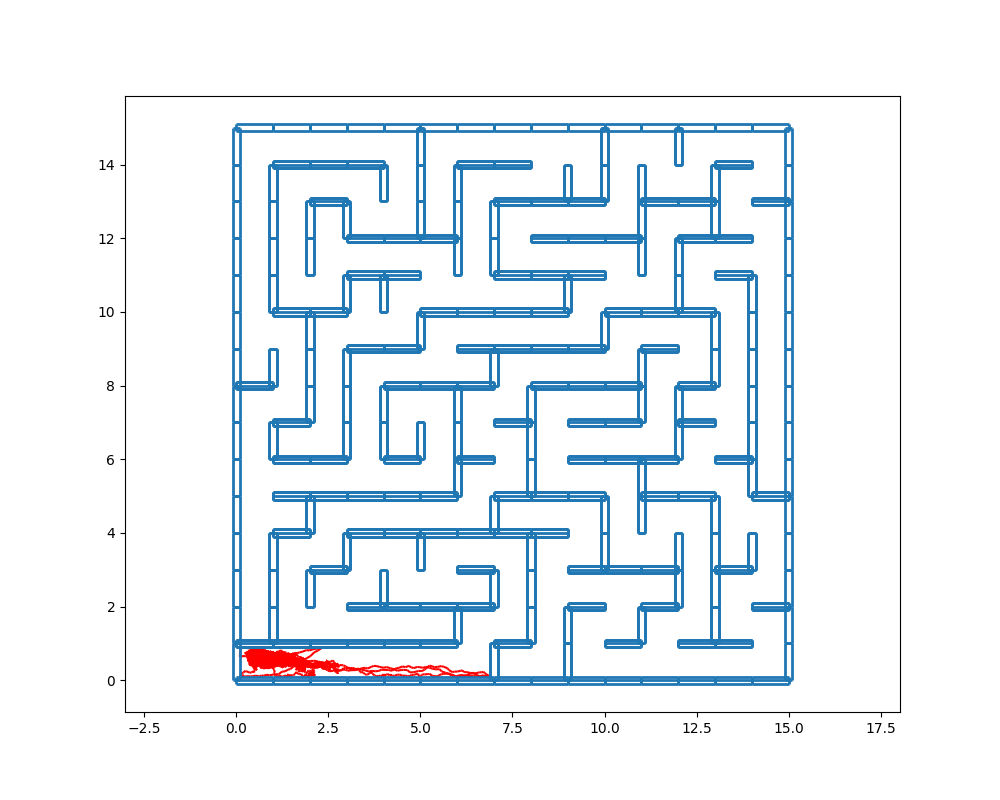} \bad{1M}& \includegraphics[trim=160px 80px 140px 80px,clip,width=2.5cm]{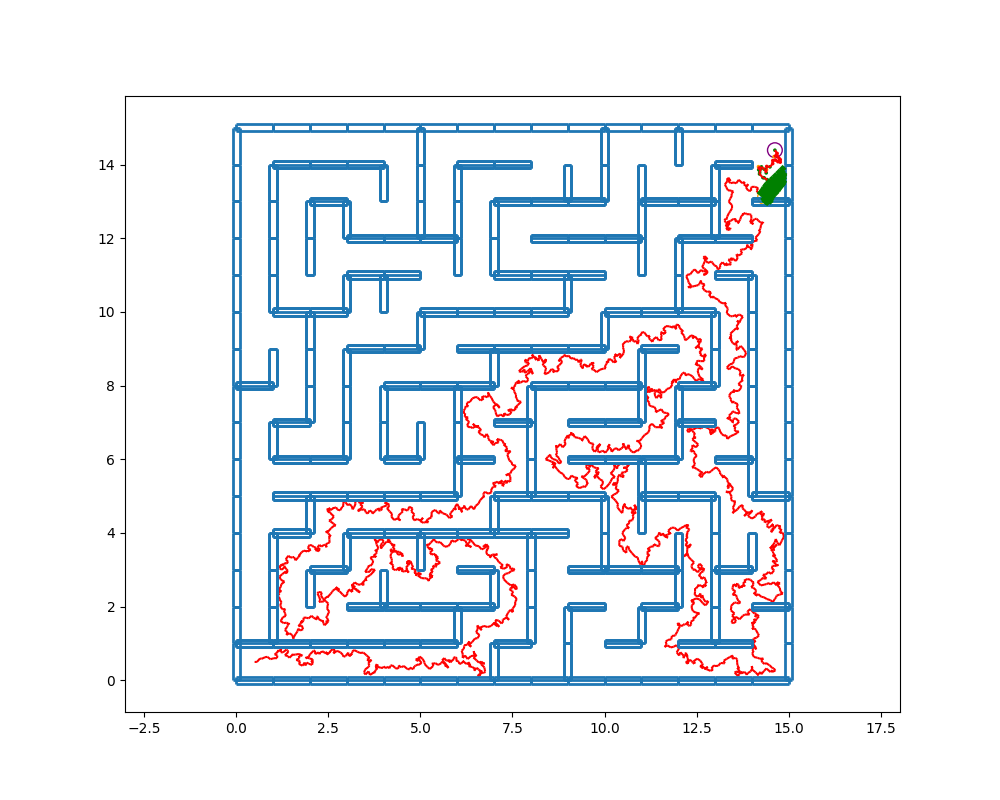} \bad{175k}& \includegraphics[trim=160px 80px 140px 80px,clip,width=2.5cm]{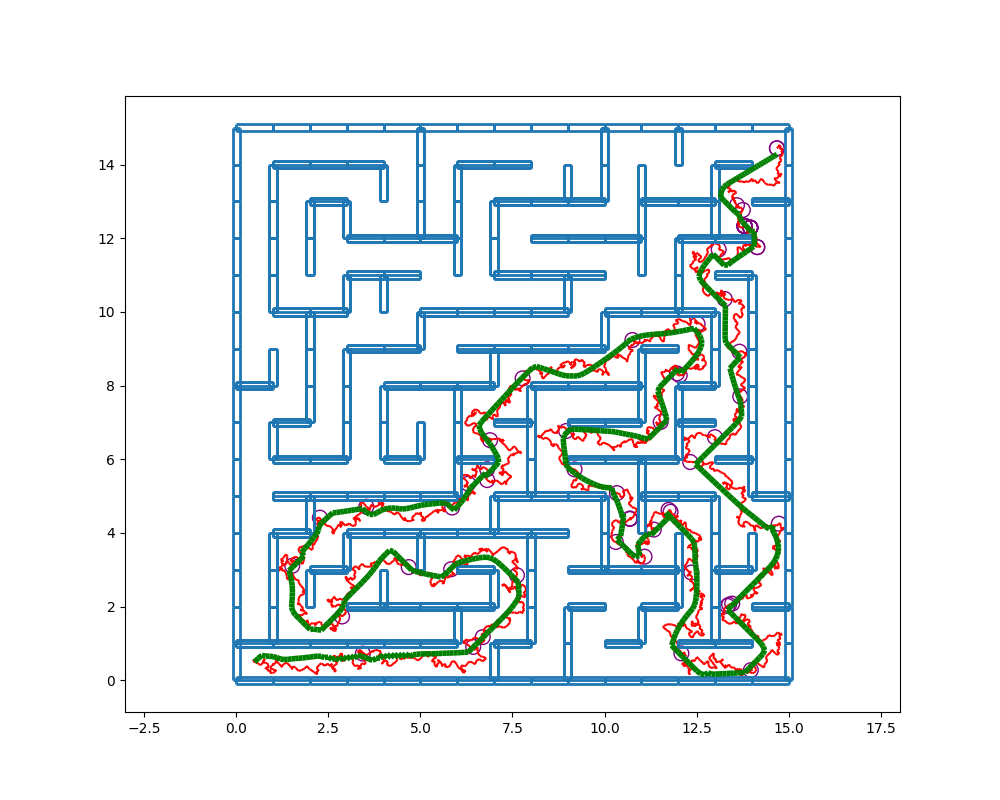} \good{22M}\\ \hline
\end{tabularx}

%% file: results_big.tex
\begin{subfigure}{.49\linewidth}
\includegraphics[trim=160px 80px 140px 80px,clip,width=\linewidth]{{results/maze05-wt0.1-s00-ddpgmpbt_ig_10}.png}
\end{subfigure}
\begin{subfigure}{.49\linewidth}
\includegraphics[trim=160px 80px 140px 80px,clip,width=\linewidth]{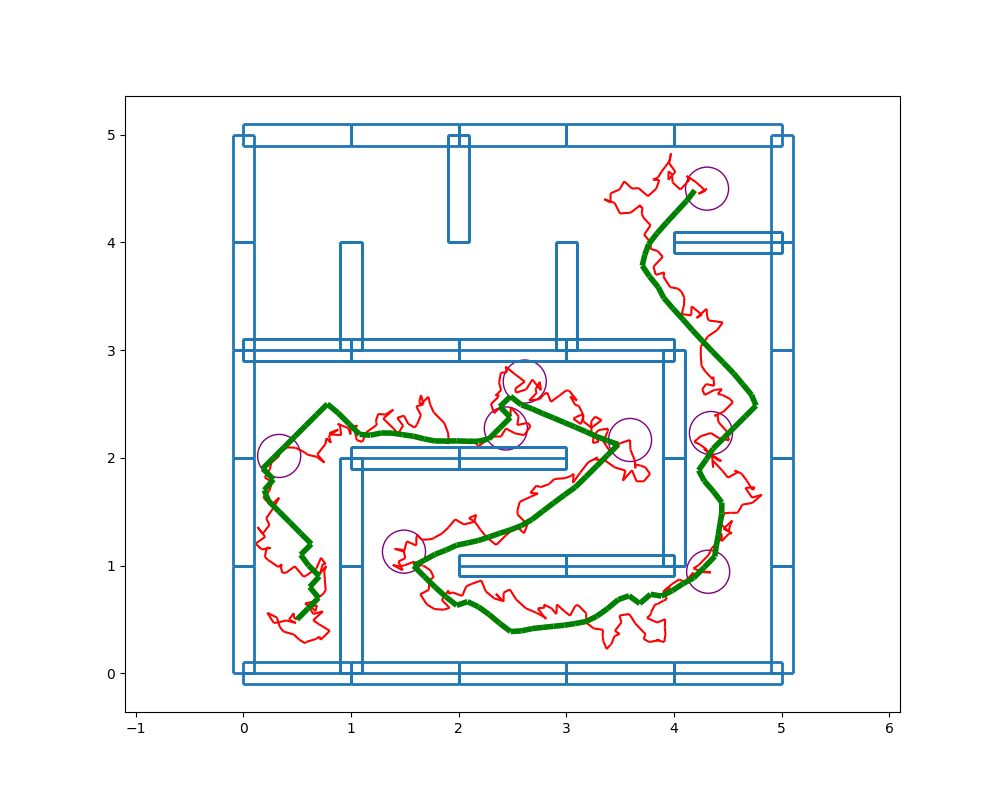}
\end{subfigure}
\begin{subfigure}{.49\linewidth}
\includegraphics[trim=160px 80px 140px 80px,clip,width=\linewidth]{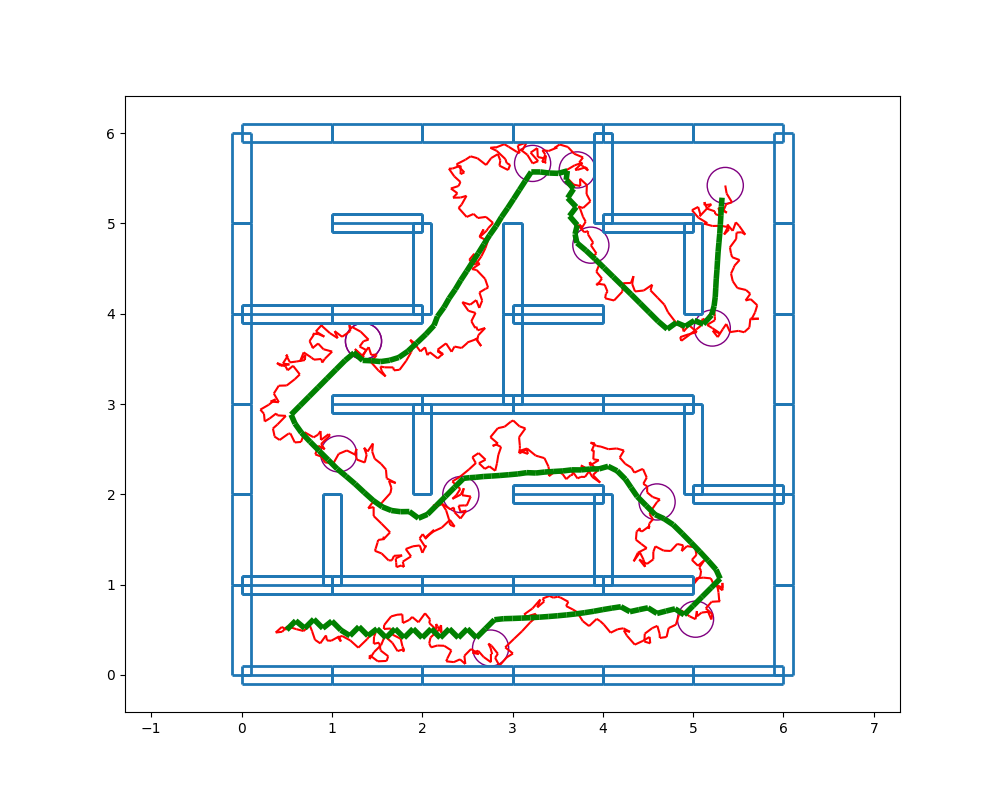}
\end{subfigure}
\begin{subfigure}{.49\linewidth}
\includegraphics[trim=160px 80px 140px 80px,clip,width=\linewidth]{{results/maze07-wt0.1-s00-ddpgmpbt_ig_16}.png}
\end{subfigure}
\begin{subfigure}{.49\linewidth}
\includegraphics[trim=160px 80px 140px 80px,clip,width=\linewidth]{{results/maze10-wt0.1-s00-ddpgmpbt_ig_31}.png}
\end{subfigure}
\begin{subfigure}{.49\linewidth}
\includegraphics[trim=160px 80px 140px 80px,clip,width=\linewidth]{{results/maze15-wt0.1-s00-ddpgmpbt_ig_66}.png}
\end{subfigure}